\documentclass[11pt]{article}
\usepackage[preprint]{acl}
\usepackage{times,latexsym}
\usepackage[T1]{fontenc}
\usepackage[utf8]{inputenc}
\usepackage{microtype,inconsolata}
\usepackage{amsmath,amssymb,booktabs,array,graphicx,xurl}
\graphicspath{{figures/}}
\newcommand{\Native}{\textsc{Native8}}
\newcommand{\Prefill}{\textsc{Prefill1}}
\newcommand{\FirstQuote}{\textsc{FirstQuote}}
\newcommand{\Full}{\textsc{Full}}
\newcommand{\code}[1]{\texttt{#1}}
\title{\microtypesetup{protrusion=false}Reading Decisions Before Spans: Efficient Generative Verification}
\author{
  GONG Yanbin\textsuperscript{1,2}\quad CHEN Lei\textsuperscript{2,3,*}
  \\[3pt]
  {\normalfont\normalsize\textsuperscript{1}ADIA} \\
  {\normalfont\normalsize\textsuperscript{2}The Hong Kong University of Science and Technology (Guangzhou)} \\
  {\normalfont\normalsize\textsuperscript{3}The Hong Kong University of Science and Technology}
  \\[3pt]
  {\normalfont\small\texttt{yanbingong@hkust-gz.edu.cn}\quad\texttt{leichen@hkust-gz.edu.cn}}
  \\[2pt]
  {\normalfont\small\textsuperscript{*}Corresponding author}
}
\date{}
\hypersetup{pdftitle={Reading Decisions Before Spans: Efficient Generative Verification},pdfauthor={GONG Yanbin and CHEN Lei},pdfkeywords={grounded verification, efficient inference, structured generation}}

\newcommand{\CenteredCascadeDevelopmentVnine}{%
\subsection{Centered routing: development and runtime isolation}
\label{app:centered-development}

\paragraph{Exploratory source-fold development.}
All measurements in this subsection use the exposed Data2txt DEV cohort:
1,026 responses in 171 whole-source families. The five-fold replay fits each
held-out fold's response threshold by training-fold F1, breaking ties toward
the largest threshold. It computes the routing radii on those training folds,
then applies the unchanged recipe to the held-out source families. The hash
controls share the calibrated fallback and match each route's actual number
of calls within its held-out fold. Table~\ref{tab:centered-oof-vnine} retains
all three prespecified budgets and both gates; the 25\% point was primary
before this measurement. The source and cached predictions were already
exposed, so this is exploratory source-fold development rather than new
independent confirmation. The replay has no new model calls or service
measurement.

\begin{table*}[t]
\centering\small
\begin{tabular}{lrrrrrr}
\toprule
DEV quantile & Centered F1 & Calls & Equal-call hash F1 & Abs0 F1 & Calls & Equal-call hash F1\\
\midrule
10\% & 91.6894 & 95 & 89.8042 & 90.1583 & 103 & 89.7436\\
25\% & 93.5172 & 250 & 90.7119 & 91.8239 & 254 & 90.9091\\
50\% & 94.5607 & 519 & 92.6762 & 94.4483 & 516 & 92.6863\\
\bottomrule
\end{tabular}
\caption{Exploratory source-fold DEV replay. Each hash column matches its
own gate's calls within folds; centered and Abs0 need not share a call count.
Always-\Prefill{} has 94.4212 F1 with 1,026 calls, and the train-fold-calibrated
encoder has 89.3847 F1 with no decoder calls. These are cached quality and
planned calls, not physical service measurements.}
\label{tab:centered-oof-vnine}
\end{table*}

At the primary point, centered routing exceeds its equal-call hash by
2.8054 F1 points (171-component conditional 95\% interval
$[1.7168,3.8944]$) and Abs0 by 1.6933 points ($[0.5498,2.8316]$), while making
250 versus Abs0's 254 planned calls. It remains 0.9040 points below
always-\Prefill{} ($[-1.6698,-0.1972]$). The 50\% point slightly exceeds the
always-generator point in this replay, but is retained as a prespecified
sensitivity point rather than chosen as the primary budget after inspection.
The intervals condition on the fixed exploratory fold predictions and do not
refit calibration within each bootstrap sample.

\paragraph{Physical deployment-fit assay.}
The deployment threshold $-0.935546875$ and centered radius $1.322265625$ are
fit once on all exposed DEV. This changes the estimation scope: the resulting
physical quality is in-sample and must not replace the source-fold result
above. An uncentered 25\% radius is 1.2578125, with the same new calibrated
fallback; it is distinct from the historical five-cutoff gate. A separately
charged label-free encoder probe fixes the source-fold hash routes before
any decoder output. Six policy arms then rotate once per original-order
response in one co-resident allocation.

The five primary runtime arms share the same single-fresh preparation for
each required model. A sixth centered arm retains the earlier preparation
path, including full/effective encoder qualification and decoder
decode--re-encode checking inside service. The fast path performs those
qualification and identity checks outside its service clock, retains their
costs, and freshly tokenizes each invoked model inside service. It does not
reuse model-input IDs. This isolates preparation overhead from a change in
the routing rule or model capability.

\begin{table}[t]
\centering\small
\begin{tabular}{lrrr}
\toprule
Full-DEV policy & F1 & 2B calls & Service (s)\\
\midrule
Always \Prefill{} & 94.4212 & 1,026 & 40.5638\\
Calibrated encoder & 90.5792 & 0 & 17.0746\\
Abs0 & 92.3960 & 257 & 26.1200\\
Centered & 94.2080 & 257 & 26.2490\\
Calibrated hash & 91.7942 & 257 & 26.2282\\
Centered, old preparation & 94.2080 & 257 & 30.2876\\
\bottomrule
\end{tabular}
\caption{Actual six-arm deployment-fit DEV assay on all 1,026 responses.
Service includes fresh required encoding, transfer, forward/generation,
readout, routing, parsing, and minimal alarm write/fsync. Shared qualification,
initialization, probe, warmup, and full diagnostics are separately paid.
These in-sample physical scores are not the source-fold estimates in
Table~\ref{tab:centered-oof-vnine}.}
\label{tab:centered-physical-vnine}
\end{table}

With 257 actual calls each, centered routing exceeds hash by 2.4138 F1
points ($[1.2859,3.5276]$) and Abs0 by 1.8120 ($[0.9020,2.7125]$).
Against always-\Prefill{}, its F1 is 0.2132 points lower
($[-0.8614,0.4301]$), while complete warm service falls 35.2897\%:
ratio 0.647103 ($[0.620627,0.673513]$). This observed small quality difference
does not establish quality equivalence. Centered routing is slightly slower
in total than Abs0 (ratio 1.004939) and matched hash (1.000791).
Against always-encoder, it gains 3.6288 F1 points ($[2.5177,4.7651]$) while
using 53.7311\% more warm service. Its precision is 94.0111\%, below Abs0's
94.8454\%, and recall is 94.4056\%, above Abs0's 90.0699\%.

The centered fast and old-preparation arms agree on all 1,026 encoder input
IDs, margins, routes, and final decisions, and on the prompt/generated IDs of
their 257 invoked decoder calls.
Changing preparation reduces service 30.2876 to 26.2490\,s, or 13.3343\%
(ratio 0.866657, interval $[0.861131,0.873117]$). This isolates runtime
overhead in this assay and supplies no new learned capability. Centered
service median is 16.749\,ms versus always-\Prefill{}'s 39.068\,ms, but its
descriptive p95 is worse: 56.924 versus 45.427\,ms. All intervals use 2,000
paired whole-family resamples and condition on the fitted deployment policy,
actual outputs, and one host/pass; they do not establish repeated-scheduling
uncertainty or production-tail behavior.

\paragraph{Complete costs and retained failures.}
All 7,184 requests complete: 6,156 measured policy requests, 1,026 probe
requests, and two warmups. They invoke 6,157 encoder forwards and 2,055 decoder
generations, with zero invalid decisions, unknown costs, or route drift.
All 2,055 actual decoder calls pass the independent CPU ID audit. Shared
decoder/encoder initialization takes 58.650/2.335\,s, within-worker native-ID
qualification 28.954\,s, asset verification 5.417\,s, label-free probe service
18.075\,s, planning 0.048\,s, and window primitives 0.172\,s. The six arms'
full-logit/ID diagnostics add 30.026\,s outside their warm service; probe and
warmup diagnostics, source preparation, final flushing, and cleanup remain
recorded. These are measured stage boundaries, not a selective per-policy
share of common cost.

The worker lasts 474.582\,s and its complete allocation costs 479 actual
GPU-seconds, counted once. Subsequent CPU ID auditing and scoring remain
outside that allocation. At this assay's closure, the complete exploration
ledger contains 114 terminal allocations and 84,690 actual GPU-seconds
(23.525 GPU-hours), including earlier failures and idle/startup work;
API expenditure remains USD~0. The 35.29\% warm-service reduction cannot be
applied to this whole-job or project total as a GPU, monetary, or energy
saving. Two local staging checks stopped before job submission; their
records remain and neither consumed a GPU allocation. All historical
uncentered results, adverse outcomes, and draft versions are preserved.
}

\begin{document}
\maketitle
\begin{abstract}
A generative verifier can flag an unsupported claim before it finishes describing the error. For applications that only screen an assistant's report, the remaining generation may be unnecessary. We examine this possibility in a frozen 2B span verifier whose output chooses between an empty error list and its first object before producing any locations. \Native{} stops at that choice. \Prefill{} supplies the seven fixed syntax tokens that precede it and generates just the unrestricted branch token, keeping the native instruction. On 2,015 code-agent TEST responses, \Prefill{} uses 84.84\% less eager resident service than full generation, with released-reference F1 of 83.43 versus 83.41. It also saves 31.98\% over \Native{} despite identical logical token totals. This additional saving remains 6.74--26.73\% under default compiled execution on the complete workload, with small changes in decisions. Evidence interventions test whether the alarm responds to the supplied facts; first-location decoding measures the additional work and localization losses when a caller also needs an error pointer.
\end{abstract}

\begin{figure}[t]
\centering
\includegraphics[width=\columnwidth]{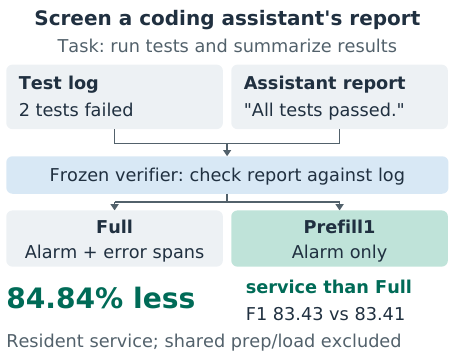}
\caption{\textbf{Screening a developer report before relying on it.} The illustrative log contradicts the assistant's report. Both artifacts reach the same frozen verifier; the application needs an alarm before inspecting error spans. Savings and F1 use all 2,015 code-agent TEST responses, with Full official as the control.}
\label{fig:teaser}
\end{figure}

\begin{figure*}[t]
\centering
\includegraphics[width=\linewidth]{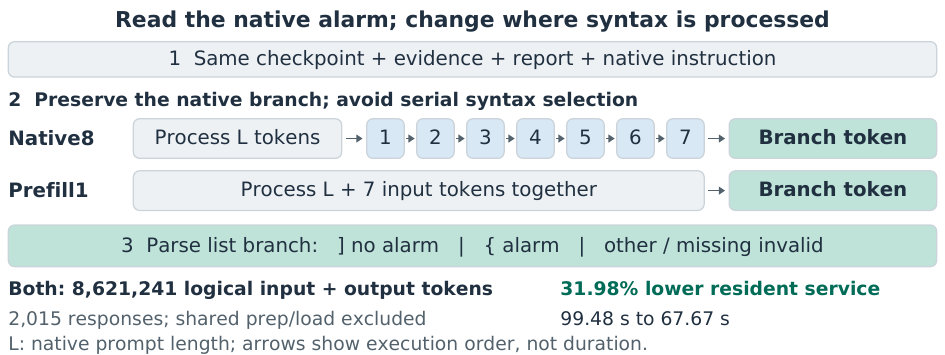}
\caption{\textbf{Two ways to read the same native branch.} \Native{} generates the fixed syntax before its decision; \Prefill{} processes that syntax as input and makes one unrestricted choice. The parser reads the reconstructed list branch in both cases. Arrows show execution order, not stage durations; service measurements count all input and output positions.}
\label{fig:method}
\end{figure*}

\section{Introduction}

A coding assistant reports ``all tests passed,'' but its execution log records two failures (Figure~\ref{fig:teaser}). The developer needs to catch this conflict before relying on the report. A generative verifier can do so by listing unsupported passages and assigning error categories. Producing the whole list, however, answers more than the initial screening question: is there anything in this report that needs inspection?

This cost is relevant wherever generated answers are checked against evidence. In RAGTruth, 43.1\% of 17,790 responses across six models contain annotated hallucinations \citep{ragtruth}. Recent benchmarks extend verification to code, developer-tool outputs, and structured documents, using detectors that generate error quotations and categories \citep{grounding}. One way to make verification cheaper is to use a compact model \citep{minicheck,lettucedetect,luna}. We consider a caller who keeps a particular verifier and wants its answer-level decision at lower cost. Asking that checkpoint for a Boolean would shorten the response, but would also change the instruction under which it decides whether the answer is supported.

We found a stopping point within the verifier's existing output format. Its response begins with a JSON error list: an empty list signals no alarm, while the first error object signals that something is unsupported. The quotation and category come later. Every output on the development source places the same seven syntax tokens before this choice. For an alarm-only request, the verifier therefore spends time generating both a known opening and an error description that the caller will not use.

The first saving is straightforward. We stop greedy generation at the eight-token branch and call this readout \Native{}. But even this short response selects seven known tokens in sequence. With \Prefill{}, we append those syntax tokens to the assistant input and ask for one unrestricted continuation. Neither a class nor a quotation appears in the supplied prefix. The checkpoint, evidence, complete answer, and native instruction stay fixed, as does the total number of input and output tokens. What changes is the schedule: output tokens are selected serially \citep{speculative}, whereas known input positions can be processed together. \Native{} gives us a direct control for the gain from this change (Figure~\ref{fig:method}).

On all 2,015 prospectively fixed code-agent TEST responses, \Prefill{} reduces eager resident service from 446.41 to 67.67 seconds (84.84\%). Its released-reference response F1 is 83.43, compared with 83.41 for full generation. The extra saving over \Native{} is 31.98\%. With default compilation and CUDA Graph execution on the complete workload, this extra saving falls to 6.74--26.73\% at sequence limits eight and one. F1 changes by $-0.12$ to $+0.27$ points across those fixed runs. The optimization thus remains useful in both configurations, though its size and some individual decisions depend on the execution path.

Assistant prefilling and skipping deterministic syntax are established techniques \citep{prefilldocs,sglang}. We use them to expose a specific decision already present in a grounded verifier: its native alarm. The study asks how much this readout saves beyond ordinary early stopping, whether it still responds to changes in the evidence, and what information is lost when generation ends there. We also continue to the first error quotation to measure the cost of recovering a location. Optimized encoder and cascade controls put these choices in context for a caller who can change models as well as readouts.

\section{Native Alarm Readouts}
\label{sec:method}

\subsection{The decision in a span report}

Given evidence $c$ and a completed answer $a$, we predict whether any answer span is unsupported according to the benchmark references. The original request is retained when it appears in the source prompt. For a verifier that returns a \code{hallucinated\_spans} list $H(c,a)$, the native alarm is
\[
\hat y_{\mathrm{raw}}=\mathbf{1}[|H(c,a)|>0].
\]
The first error object makes this list nonempty, regardless of which quotations or categories follow. All readouts receive the same source prompt/context, completed answer, and opaque identifier. Gold spans, categories, and generator metadata are used only by the evaluator.

We evaluate full generation in two ways. \emph{Full raw} reads list presence directly. \emph{Full official} uses the public evaluator's quotation-to-answer-offset mapping \citep{officialeval}; a nonempty list can become a negative result if none of its quotations maps to a valid location. Both readers use exactly the same generated output and incur the same generation cost. Keeping them separate lets us trace a disagreement either to the native list choice or to the localization step applied after generation.

\subsection{Stopping and prefilling at the branch}

Ordinary generation on the exposed development source begins with a fixed seven-token prefix $s$. For this checkpoint, $s$ decodes to \code{\{"hallucinated\_spans":} and stops just before the list-opening choice. \Native{} uses the native prompt and greedy decoder with an eight-token output cap. Its parser verifies the prefix and reads the branch; it need not wait for a quotation or a complete JSON response. If the branch is missing or unexpected, the result is invalid. We retain that result without a repair call or fallback decision.

For \Prefill{}, we append the seven IDs of $s$ to the tokenized native assistant prompt. The model then generates one greedy token $z$ from the unrestricted vocabulary. We decode $s\mathbin{\|}z$, find the first JSON object, and inspect the opening of its \code{hallucinated\_spans} list. The next non-whitespace character determines the result: \code{]} means no alarm and \code{\{} means an alarm. A missing branch or any other character is invalid. This procedure supplies only the fixed syntax; the model must still choose the branch. We provide no candidate quotation, class choice, or vocabulary mask (Figure~\ref{fig:method}).

We apply that same parser to the eight tokens generated by \Native{} and retain the actual input/output IDs for both readouts, including invalid results. Appendix~\ref{app:protocol} gives the pinned prefix IDs, decoding options, and validity checks. Prefix selection uses DEV only; no test output is used to choose a prefix, mask, or threshold. To check the effect of retaining the native format, Section~\ref{sec:mechanism} also compares a semantics-matched binary output contract at its fixed default operating point.

\subsection{Why the execution cost can change}

Let $P(L)$ be prompt processing through the first next-token logits and $D(L+i)$ a subsequent decode step. Ignoring bookkeeping, the two short readouts cost
\begin{align}
C_8(L)&\simeq P(L)+\sum_{i=1}^{7}D(L+i),\\
C_{\mathrm{prefill1}}(L)&\simeq P(L+7).
\end{align}
In \Native{}, each selected output token becomes input to the next decode step. \Prefill{} already knows all seven syntax positions, so it processes them with the prompt. Both readouts still receive the complete evidence and answer. Counting the supplied syntax as input makes their logical token totals equal, despite the difference in execution order.

This model predicts a saving when $P(L+7)-P(L)<\sum_{i=1}^{7}D(L+i)$, or when processing the extra prompt positions costs less than the decode steps they replace. Token counts alone cannot tell us whether the condition holds. Batching can amortize decode work; with long inputs, prompt processing can dominate both expressions. In that case, even a substantial reduction in the cost of the seven syntax positions has little effect on total service. We therefore compare the readouts under matched scheduling and repeat the measurement with compiled execution.

\subsection{When the branch is preserved}

Suppose ordinary greedy generation emits $s$. With identical causal computation, arithmetic, and tie-breaking, supplying that same $s$ as input gives the same next-token branch. This conditional identity motivates our use of the class-independent syntax and an unrestricted continuation. It applies to the native list choice. An alarm may still be suppressed later by the official mapper if the generated quotation has no valid answer offset.

Actual prefill and decode calls can use different kernels and batch shapes, changing their finite-precision trajectories. If the all-vocabulary logits at the completed prefix differ by at most $\epsilon$ per coordinate, a top-token gap greater than $2\epsilon$ is sufficient to keep the same winner: the leading logit can fall by $\epsilon$ while a competitor rises by $\epsilon$. We retain selected IDs rather than complete logit vectors, so our traces cannot certify this margin. Instead, we measure prefix and branch agreement and report paired F1 changes. Appendix~\ref{app:native_branch_theory} gives the conditional identity proof and bounds F1 changes in terms of changed decisions.

\section{Experimental Design}
\label{sec:eval}

We use full generation to measure the cost of the complete report and optimized early stopping to isolate the additional benefit of prefilling. Comparing the raw and official readers distinguishes native alarms from localization postprocessing. Binary-contract and evidence-flip controls then test how the decision depends on its output interface and the supplied facts. Finally, encoder and cascade comparisons examine a different choice available to the caller: allocating work across models with different inherited detection capabilities.

\subsection{Models and comparison points}
The models are the public LettuceDetect Qwen3.5-2B span verifier and its same-training 307M mmBERT-base counterpart \citep{grounding,mmbert}. We compare full official/raw output, \Native{}, and the encoder at native and DEV-calibrated thresholds. For the encoder, an optimized FA3-window implementation retains the same checkpoint, BF16 parameters, FP32 rotary buffers, and fixed thresholds. This optimization reverses the earlier cross-model speed ordering, making the encoder the faster option. Full output is capped at 1,024 tokens, and all native decoder readouts use unrestricted greedy generation.

\begin{figure*}[t]
\centering
\includegraphics[width=\linewidth]{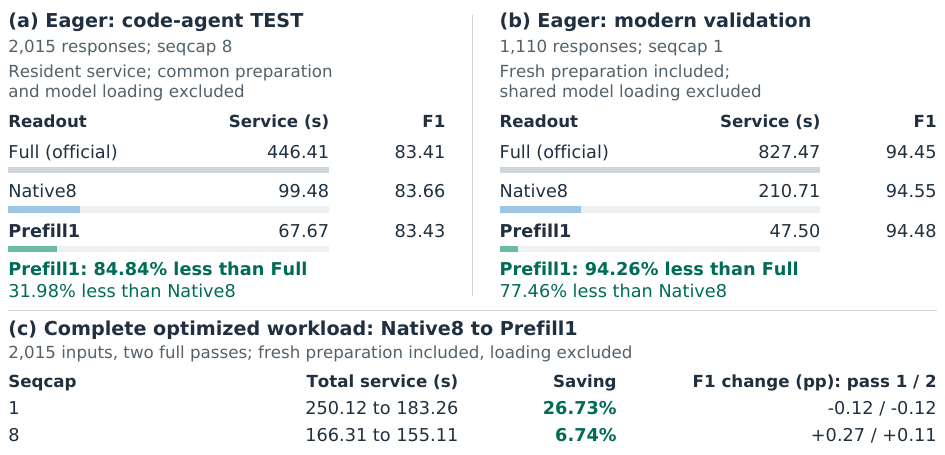}
\caption{\textbf{Quality and service cost of native readouts.} Top: eager execution on two cohorts, with separate bar scales and clocks. Bottom: default compiled execution on the complete code-agent workload; seqcap is the maximum active sequences, and service totals include two full passes. F1 measures agreement with released references. Three of 126 paired observations at seqcap eight are slower with \Prefill{}. Paired intervals and decision changes are reported in the text.}
\label{fig:gains}
\end{figure*}

\subsection{Evaluation populations and references}
Development uses an already exposed Data2txt source from upstream TRAIN: 1,026 responses in 171 families. The main confirmation covers \emph{all} 2,015 code-agent TEST responses, including 1,014 reference positives in 1,613 source-derived components. Methods, selection, and measurement are fixed before opening this source. Two further fixed sources contain 617 tool-output and 641 README responses; \Prefill{} is not run on the earlier tool source. Quality, class, length, and outcome never determine whether a selected response is retained.

For a later comparison, eleven policies are fixed before opening a project-unseen modern validation cohort of 1,117 code-agent, tool-output, and README responses. Source-only closure links seven responses to previously evaluated inputs. We execute all of them, with the remaining 1,110 responses/639 components forming the predeclared primary population; the appendix also reports all 1,117 and each source separately. This is \emph{upstream validation}, used to select the encoder checkpoint; the generator's selection rule is unspecified. It extends the workload and cost comparison without replacing the principal TEST evidence.

The public benchmark contains localized edits and released span annotations \citep{grounding,dataset}. Its code-agent TEST references include the data authors' model-assisted triage and review. README instead uses lower-precision first-pass references, and some of its edits are detectable without context. Both models trained on unified TRAIN, so these experiments compare methods within the task against released references. Long contexts and realistic label noise affect hallucination-detector evaluation \citep{rethinkeval}; we report each source's provenance and complete selected population.

All native readouts receive the same evidence, answers, and instructions. The encoder accepts 8,192 tokens and the decoder 32,768; encoder truncation removes context only. The main analysis retains all 441 truncated code-agent contexts, accompanied by a prospectively fixed no-truncation subset of 1,574 responses. Source components join exact normalized questions/contexts and context near-duplicates with five-word Jaccard similarity $\geq0.85$. These groups account for dependence in the summaries, although they do not establish repository independence. Appendix~\ref{app:protocol} records the full protocol and its revisions.

\subsection{Quality and cost measurements}
Positive-class response F1 is the primary quality metric. Paired 95\% percentile intervals use 2,000 whole-component bootstrap resamples shared across contrasts, conditional on the fixed outputs and grouping. Before scoring against references, we audit the actual samplers, token IDs, offsets, and request counts. Native readouts and the grammar-complete binary control produce no invalid decisions. For other controls, invalid outputs remain in the record with sensitivity analyses; incomplete runs receive no partial-cohort score.

We measure cost at two service boundaries. The principal experiment uses a resident decoder on one H100 80GB in BF16/eager mode, without prefix caching and with at most eight active sequences. Original-order chunks of 32 balance the order of the three arms. Its \emph{resident service} timer includes generation, parsing, and result handling; shared initialization and source preparation are excluded.

The later eleven-arm experiment uses one decoder sequence and a co-resident optimized encoder. Each arm pays for \emph{fresh service}, including required tokenization, transfer, model calls, readout/routing, parsing, and the same minimal alarm write/fsync. Arms rotate per response. Shared initialization, qualification, probes, and full diagnostics are recorded separately. Figure~\ref{fig:gains} separates the experiments, whose totals cannot support cross-workload latency comparisons. Main-job GPU accounting includes startup, failures, and idle/CPU time, each counted once.

For each comparison, we compute savings from the ratio of complete-cohort service totals. This weights each request by its measured cost; averaging per-response speedups would answer a different question. We keep workload results separate because context length, scheduling, and preparation requirements all affect these totals.

\section{Verification quality and service cost}
\label{sec:results}

\subsection{Primary test: 84.84\% less service}

Across all 2,015 code-agent responses, full official, \Native{}, and \Prefill{} obtain 83.4128, 83.6594, and 83.4304 F1, respectively, using 446.4075, 99.4773, and 67.6682 seconds of service. Stopping with \Native{} removes 77.72\% of full service. \Prefill{} increases this reduction to \textbf{84.84\%}, saving a further \textbf{31.98\%} over the eight-token control (Figure~\ref{fig:gains}, left).

\begin{table}[t]
\centering\small
\begin{tabular}{lrr}
\toprule
\Prefill{} F1 contrast & $\Delta$ (pp) & 95\% interval\\
\midrule
Code-agent: full official & +0.018 & $[-0.273,+0.327]$\\
Code-agent: \Native{} & $-$0.229 & $[-0.548,+0.059]$\\
Modern: full official & +0.020 & $[-0.336,+0.401]$\\
Modern: \Native{}/full raw & $-$0.077 & $[-0.363,+0.212]$\\
\bottomrule
\end{tabular}
\caption{Paired quality contrasts for the two eager experiments in Figure~\ref{fig:gains}. Intervals use whole-source components and retain possible losses. Full raw on code-agent has 83.5363 F1 from the same full generation.}
\label{tab:primary}
\end{table}

The close F1 scores do not mean the readouts make identical decisions. \Prefill{} yields 788/87/226/914 TP/FP/FN/TN, against 787/86/227/915 for full official. In the token audit, all full and eight-token outputs begin with the same seven syntax tokens, yet their first eight IDs differ on eight responses. Appending each prefill output to the supplied prefix also reveals eight differences from the eight-token outputs. These disagreements occur at the decision after the shared syntax.

Postprocessing accounts for a further difference. The official reader removes two positive full-output alarms when their quotations fail to map to answer offsets. Reading the same output with the raw reader gives 83.5363 F1. The early readout targets this native list choice, while the official score includes the later mapping step. Table~\ref{tab:primary} reports the paired quality differences and their uncertainty.

Full generation emits 95,134 tokens; \Native{} emits 16,120 and \Prefill{} emits 2,015. The shorter output in \Prefill{} is partly offset by 14,105 additional input tokens. Once those are charged, the two short readouts each process exactly \textbf{8,621,241 logical input-plus-output tokens}. Their 31.98\% service difference remains even though the totals match. Compared with full generation, the reduction in logical token volume is only 0.91\%. The service savings should therefore not be read as corresponding reductions in API prices or allocated-GPU charges.

\subsection{Additional workloads and fresh preparation}
\label{sec:modern-confirmation}

On the separately frozen README source, all four decoder decisions agree on each of the 641 responses, yielding 98.5782 F1. \Prefill{} takes 8.3065 seconds, compared with 85.2763 for full generation and 22.1382 for \Native{}, saving 90.26\% and 62.48\%, respectively. On the earlier 617-response tool source, full raw and \Native{} also agree throughout at 90.9091 F1; \Native{} reduces total worker time by 46.83\%, with loading reported separately. These comparisons retain the source-specific references and timer boundaries described in Section~\ref{sec:eval}.

Fresh preparation is included for every policy in the later 1,110-response comparison (Figure~\ref{fig:gains}, right). Here \Prefill{} obtains 94.4751 F1, compared with 94.4547 for full official and 94.5522 for \Native{}. It reduces service from 827.4728 to 47.4990 seconds relative to full generation, and from 210.7107 relative to \Native{}: savings of \textbf{94.26\%} and \textbf{77.46\%}. The paired service ratios are 0.057403 ($[0.053997,0.061365]$) and 0.225423 ($[0.218730,0.234101]$). Full raw and \Native{} have the same aggregate F1; Table~\ref{tab:primary} gives the paired quality uncertainty.

Here each policy uses a single active sequence and pays for fresh preparation. The original batch experiment uses a different cost boundary, so its smaller savings cannot be attributed to the workload alone. Appendix~\ref{app:modern-confirmation-vnine} gives all eleven policies, separate results for the three sources, the no-truncation analyses, and results for the full cohort.

\subsection{Does the saving survive runtime optimization?}

The eager measurements leave a practical question: does prefilling still help when the engine optimizes decoding? We replay all 2,015 fixed inputs with default compilation and CUDA Graphs on one H100. Each of the 63 original-order chunks runs both arms in AB/BA order, with the initial order counterbalanced by chunk. We use sequence limits one/eight and disable prefix caching. Fresh service counts tokenization, generation, parsing, and a minimal result write.

Across both passes, \Native{} and \Prefill{} require 250.12/183.26 seconds at limit one and 166.31/155.11 at limit eight, saving \textbf{26.73\%}/\textbf{6.74\%} at identical logical token totals. The saving is smaller at greater concurrency, and three of 126 paired batch-order observations at limit eight favor \Native{}. The arms also differ on 12/4,030 and 13/4,030 branches. Joining these outputs to existing references gives descriptive F1 changes of $-0.12$ in both limit-one passes and $+0.27$/$+0.11$ at limit eight; these differences rule out lossless decoding, and the reused references provide no independent quality confirmation. Initialization is timed separately. Appendix~\ref{app:optimized_full_vseventeen} reports every clock and quality result, with the earlier 24-input pilot retained in Appendix~\ref{app:optimized_runtime_vfifteen}.

\paragraph{Complete accounting.}
The primary allocation uses 1,296 GPU-seconds for initialization, preparation, and decoder/encoder workers; the eleven-arm allocation uses 1,907. The earlier compiled-execution and binary-contract controls use 744 GPU-seconds, and the full-workload optimized run adds 1,697. We account for common work separately rather than assigning it selectively to one arm. Including all failed and exploratory allocations, total exploration uses 90,659 GPU-seconds across 121 terminal allocations. This total includes 747 seconds of auxiliary analysis outside the paper and is separate from the method service timings.

\section{Evidence dependence and output requirements}
\label{sec:mechanism}

\subsection{Interface fidelity and evidence response}

Changing the output contract to a binary JSON field leaves the verification definitions, categories, evidence, and answer unchanged. To make this control grammar-complete, its one-token mask admits all six vocabulary spellings of \code{true}/\code{false} with legal JSON whitespace, selected by grammar rather than labels. Across all 1,026 exposed DEV responses/171 source families, native \Prefill{} obtains 94.36 F1 and this control 92.04, a difference of $+2.32$ points ($[1.10,3.59]$), at comparable fresh service (14.38/14.66\,s). Both produce valid decisions throughout. The native precision/recall pair is 94.03/94.69\%, versus 96.33/88.11\% for binary: under these defaults, native readout trades precision for recall. Unrestricted binary generation produces 659 invalid outputs and is therefore not the strong baseline. Appendix~\ref{app:binary_contract_vfifteen} preserves it alongside the earlier Yes/No recipe and incomplete-mask results. This fixed-default DEV comparison does not establish intrinsic quality superiority over score-aggregated or calibrated binary readouts.

Agreement with full generation alone would not show that an alarm uses the evidence. We construct 96 inputs from 24 frozen tool-return families by crossing two recorded facts with two fixed literal answers. Full official/raw, \Native{}, and \Prefill{} solve all 24 quartets and all 48 fixed-answer evidence-flip pairs. Each answer and each fact occurs under both label conditions, which bounds a deterministic answer-only or context-only classifier at 50\% accuracy. When we change the recorded fact but keep the answer, the models give the correct opposite alarm. This result tests conditionality on literal records; it says nothing about whether the recorded execution actually succeeded (Appendix~\ref{app:native_validity}).

\subsection{Restoring a first location costs more}

To obtain an error pointer, we continue decoding past the alarm. \FirstQuote{} supplies the same syntax and stops at the marker closing the first predicted quotation. The official mapper attempts to turn that quotation into a character location. If it cannot, we retain an alarm without a location; neither repair nor another model call is allowed. The matched full control uses the same prefix/detokenization and selects its first valid quotation after completing generation.

On the already evaluated code-agent cohort, \FirstQuote{} retains every observed raw full alarm and its 83.5363 F1. Matched service falls from 425.791 to 204.637 seconds, a reduction of 51.94\%. Some locations are lost: valid-location coverage among predicted alarms drops from 874/875 to 861/875, and offset-presence F1 falls by 0.5946 points ($[-1.0496,-0.2048]$). A first quotation thus supplies more information than an alarm at a measured additional cost, but it does not recover all the locations found by completing the report.

\subsection{A fast encoder and selective escalation}
\label{sec:centered-cascade}

The optimized encoder provides a cheaper model-level alternative to shortening the generator's output. On primary code-agent TEST, its fixed-threshold F1 is 77.3345 against 83.4304 for \Prefill{}, a 6.0958-point difference ($[4.0262,8.2156]$). The difference remains 6.0398 points on the no-truncation subset and reflects inherited checkpoint capability, rather than a quality gain from prefilling. The encoder nevertheless has higher recall, 84.12\% versus 77.71\%. A fixed 18-source timing control with three repetitions shows that legal FA3 windows reduce its median service from 276.547 to 44.121\,ms, reversing the earlier warm speed ordering. The generator readout offers higher F1 in this comparison, while the optimized encoder offers lower service cost.

A physical cascade combines these options by running the encoder on every response and escalating uncertain cases to \Prefill{}. Let $m$ be the maximum answer-token logit difference, with calibrated decision $\mathbf{1}[m>t]$. Centered routing escalates when $|m-t|\leq r_q$; the simpler Abs0 control uses $|m|$ and the same calibrated fallback. Five source-fold DEV estimates precede one all-DEV fit, fixing the threshold at $-0.935546875$. The DEV quantiles are $q\in\{.10,.25,.50\}$, with $.25$ primary. Routes and thresholds remain fixed on modern validation. Equal-call hash routing measures the value of selection at each budget, and each route pays for fresh encoder processing plus any invoked decoder processing.

\begin{table}[t]
\centering\small
\begin{tabular}{lrrr}
\toprule
Modern validation policy & F1 & 2B calls & Service (s)\\
\midrule
Always \Prefill{} & 94.4751 & 1,110 & 47.4990\\
Calibrated encoder & 85.0168 & 0 & 23.1406\\
Abs0, DEV q25 & 91.0268 & 216 & 31.5589\\
Centered, DEV q25 & 90.3795 & 250 & 34.1989\\
Matched hash, q25 & 86.7761 & 250 & 32.9830\\
\bottomrule
\end{tabular}
\caption{Alternative working points under the same fresh-service boundary on 1,110 modern validation responses. All models required by each policy are charged. Native encoder F1 is 85.5072 from the same forward pass. All three predefined routing budgets remain in the appendix.}
\label{tab:modern-vnine}
\end{table}

Centered selection gains 3.6034 F1 points over exact-250-call hash routing ($[2.1127,5.2207]$) but uses 3.69\% more service, since equal call counts can still select different input lengths. It saves 28.00\% service over always-\Prefill{} while losing 4.0956 F1 points. The centered rule's DEV advantage over Abs0 does not carry over: Abs0 obtains 91.0268 F1 with 216 calls and 31.5589 seconds, compared with 90.3795, 250, and 34.1989 for centered routing. Centered q10/q50 yield 88.0069/93.3937 F1 and 26.4375/40.3954 seconds; we retain the predefined primary budget rather than replacing it with q50. Thus the simpler rule is preferable on this cohort, even though selection improves on random allocation at matched calls. Appendices~\ref{app:centered-development} and~\ref{app:modern-confirmation-vnine} report development, the complete quality--cost frontier, and the adverse tail-latency results.

\subsection{Choosing what verification should return}

A developer dashboard can send a completed report and the logs or documents available to the assistant to the verifier, then flag reports for inspection. For example, a claim that all tests passed can be flagged against a supplied failure log without enumerating every unsupported quotation. If the dashboard instead highlights the offending words, it needs locations; showing all spans and categories requires the full report. In each case, the flag guides inspection. It does not require automatic rejection of flagged reports or acceptance of unflagged ones.

The caller's requested output determines which costs can be compared. An alarm cannot locate or repair an error. Continuing to a first quotation recovers some pointers, and our experiment records both its extra work and the locations it misses. A later localization pass must also be charged if the application requests it. Choosing a faster detector is a separate option: the optimized encoder has lower service cost, and a cascade adds generator calls to selected requests. Native readouts instead let the caller vary how much output it reads from a fixed verifier.

\section{Related Work}

\paragraph{Grounded verification at different granularities.}
RAGTruth annotates hallucinations at word level in naturally generated responses \citep{ragtruth}. MiniCheck studies efficient document-grounded fact checking \citep{minicheck}, while LettuceDetect and Luna use compact encoders to localize unsupported content \citep{lettucedetect,luna}. The code, tool-output, and structured-document evaluation of \citet{grounding} supplies our benchmark and both checkpoints. It already reports span- and example-level F1. We use those same references to compare the output read from a fixed verifier, measuring its service cost and the functionality retained. Scores from other source mixtures do not provide head-to-head comparisons for this question.

\paragraph{Early verification and short decisions.}
PrefixNLI trains entailment models on incomplete \emph{answer} prefixes to provide feedback during generation \citep{prefixnli}. Our answer is complete before verification starts; we shorten the \emph{verifier's output}. \citet{firsttoken} estimates correctness confidence from first-content-token entropy in closed-book factual QA. Here the decision comes from the learned structural branch of a context-grounded detector. Reasoning's Razor compares reasoning modes with classification-prefilled prompts and normalized label-token scores \citep{razor}. Our prefix preserves the detector's native span format, and the decision is an unrestricted greedy continuation at its list branch.

\paragraph{Structured decoding and model allocation.}
Assistant continuation makes prefix input possible \citep{prefilldocs}. SGLang's compressed finite-state machines use it to jump over deterministic portions of structured output \citep{sglang}. We build on this syntax-skipping principle by identifying where a native span detector has already supplied an alarm. The experiments measure its increment over ordinary early stopping, with official/raw full readouts and evidence and localization controls. The auxiliary routing study follows confidence-based model allocation in FrugalGPT and AutoMix \citep{frugalgpt,automix}. It combines the optimized encoder with a native-alarm call and records the resulting service cost, including the simpler gate's stronger transfer.

\section{Conclusion}

The frozen span verifier signals an error before describing it. Stopping at this branch removes the localization work, while supplying its fixed syntax as input removes serial formatting decisions. Both readouts keep the native instruction. On code-agent TEST, \Prefill{} uses 84.84\% less eager resident service than full generation and 31.98\% less than eight-token stopping. Under default compiled execution on the complete workload, the extra saving is 6.74--26.73\%, with F1 changes between $-0.12$ and $+0.27$ points. Early stopping accounts for most of the saving. The comparison with \Native{} isolates prefilling's smaller, runtime-dependent gain, which full-output comparisons would conflate with early stopping. Execution can also change individual decisions.

When a caller needs a location, first-quotation decoding saves 51.94\% of matched service and retains all observed raw alarms, at a loss of 0.59 offset F1 points. Completing generation provides the remaining descriptions. The required output, whether an alarm, a first pointer, or a complete report, determines where decoding should stop.

\section*{Limitations}

We study one frozen 2B checkpoint with a specified native grammar. The conditional identity assumes that ordinary generation emits the supplied prefix under identical computation. Kernels, batch shapes, and finite precision can violate that computational match, and the empirical comparisons report the resulting decision changes. Bootstrap intervals describe uncertainty conditional on the fixed outputs and source grouping; they do not establish population equivalence.

Quality is measured against released span references whose precision varies by source. Both models trained on unified TRAIN. DEV was exposed, and later functional/runtime controls reuse closed inputs. The additional cohort is upstream validation used to select the encoder checkpoint; the generator's selection rule is unspecified. Lexical closure reduces overlap but does not certify semantic or repository independence, or rule out contamination during model training.

An alarm provides neither locations nor categories, and the first quotation gives incomplete localization. The evidence intervention uses constructed literal records. These tests concern the relation between a supplied record and a report, so a successful alarm does not certify program execution, patch success, or every claim in a deployed report. The results apply to the verification requests evaluated here.

Service cost depends on the workload, engine, batching, and measurement boundary. The original batch measurement excludes shared preparation; the later fresh-service measurement includes it. The compiled comparison uses the default optimized execution path and therefore cannot isolate the effect of CUDA Graphs alone. It covers 63 fixed chunks in both orders on one allocation, including reverse service comparisons and small decision changes; startup and capture costs are recorded separately. Batch service, per-request latency, and allocation billing measure different costs. The optimized encoder is faster, and cross-model quality differences also reflect checkpoint capability. Finally, the binary comparison uses one fixed contract on exposed DEV. It is neither a new modern TEST result nor evidence that span training is necessary.

\section*{Ethical Considerations}

This work uses released data and frozen models. Binary alarms can miss unsupported content or flag supported text and should assist inspection, not certify safety or correctness. Dataset/model rights, sensitive content, and intended artifact licenses require human-author review before external release.

\section*{Acknowledgements}

OpenAI Codex assisted with experiment orchestration, analysis, figure preparation. AI tools are not authors. The human authors are responsible for the content, accuracy, attribution, and final approval of this work.

\bibliography{references}
\clearpage
\appendix
\section{Complete data and execution protocol}
\label{app:protocol}

\subsection{How to reproduce, audit, and adapt the readout}
\label{app:reading-guide}

The appendix follows the decisions needed to implement and inspect the method.
Section~\ref{app:protocol} fixes the input interface, tokenizer/model revisions,
prefix, parsing rules, data boundaries, and cost scopes.
Section~\ref{app:measurements} retains the complete source comparisons,
localization measurements, binary-prompt control, and encoder implementations.
Section~\ref{app:cascade} supplies the optional routing development and all
frozen transfer points. The conditional argument and decision-change accounting
are in Section~\ref{app:native_branch_theory}; evidence interventions and all
runtime orders are in Section~\ref{app:native_validity}.
Section~\ref{app:increment_controls_vfifteen} adds the semantics-matched binary
contract and compiled-execution controls, with their full grammar and cost audits.
Section~\ref{app:artifacts} connects these results to their preserved evidence
and complete exploration costs.

To reconstruct the focal inference path, tokenize the unchanged native request
containing the evidence and complete answer. For \Native{}, generate at most
eight tokens from that prompt. For \Prefill{}, append the pinned seven syntax
\emph{token IDs} to the prompt IDs and generate at most one unrestricted greedy
token. Decode the generated IDs for \Native{}, or the concatenation of supplied
and generated IDs for \Prefill{}. At the first JSON object's
\code{hallucinated\_spans} list, the whitespace-tolerant parser maps
\code{]} to no alarm and \code{\{} to an alarm. A missing or unexpected branch,
an empty generation, an exceeded cap, or an unexpected termination remains
invalid. The short-output parser does not require completed JSON, and neither
readout applies the later quotation-to-offset map. Actual prompt and output IDs
are audited independently of this text parser. Decoding skips special tokens and
disables tokenization-space cleanup; accepted termination reasons are stop and
length. The native request is tokenized once, and prefix IDs are appended without
retokenizing the decoded concatenation.

The manuscript source archive contains the text, bibliography, and figures
needed to compile the paper. Reproducibility artifacts are maintained
separately. The portable core implements full native generation, \Native{},
and \Prefill{}, with pinned model and dataset identities, a source adapter,
and a CPU scorer. Aggregate CSV tables record the primary readouts, cascades,
branch stability, evidence interventions, runtime batches, and interface
controls. JSON manifests preserve worker/scorer hashes, protocol identities,
corrections, and complete exploration costs. Offline checks reconstruct the
primary F1 scores, optimized service totals, and allocation costs from those
aggregates. These checks are not a fresh GPU reproduction, and the portable
core does not reimplement the auxiliary controls. Model weights, raw benchmark
text, evaluator references, and original execution traces are not included
in this manuscript-only archive; independent rerunning requires the
corresponding pinned assets and execution records.

Adapting the idea to another verifier requires a new native-interface check,
not reuse of the number seven. On development data, identify a class-independent
prefix followed by a branch that determines the requested output. A rationale
or other input-dependent content before that branch cannot simply be supplied
as fixed syntax. Freeze the new prefix, parser, caps, and cost scope before
comparison on separate inputs; compare full raw, full postprocessed, native
early stopping, and supplied-prefix continuation. Preserve invalid branches,
all input/output token costs, and any functionality lost by stopping. This
procedure describes how to test an extension; the present experiments establish
the result for the specified frozen verifier.

\subsection{Frozen prefix and routing parameters}

For the pinned decoder tokenizer, the seven class-independent syntax IDs are
(4754,\allowbreak 40839,\allowbreak 1692,\allowbreak 15025,\allowbreak 9823,\allowbreak 581,\allowbreak 763). The branch follows this exact sequence;
the prefix does not include a class-dependent list item or empty-list choice.
The fixed native parser accepts an empty-list or first-object branch and retains
other continuations as invalid. Actual input/output ID audits check this native
interface separately from the quotation-to-offset mapping of complete outputs.

For the later centered cascade, a single full-DEV deployment fit fixes
$t=-0.935546875$ and radii $r_{.10}=0.541015625$,
$r_{.25}=1.322265625$, and $r_{.50}=2.384765625$. Each radius is the sorted
DEV distance at zero-based index $\lfloor q(n-1)\rfloor$; all ties escalate.
The q25 radius is the primary point. These are development quantiles, not a
promise of identical call fractions after transfer.

Let $C_E(x)$ and $C_D(x)$ include each model's fresh preparation and service,
and let $C_R(x)$ denote routing overhead. For gate $g(x)$ the request cost is
$C_g(x)=C_E(x)+g(x)C_D(x)+C_R(x)$. Relative to always using the decoder,
expected warm cost falls only when
\[
\mathbb{E}[(1-g(x))C_D(x)]>
\mathbb{E}[C_E(x)+C_R(x)].
\]
The condition permits input length and routing to be correlated: call fraction
alone cannot determine savings. Shared initialization and qualification are
separately charged.

\subsection{Development and test boundaries}

Table~\ref{tab:cohorts} distinguishes the initial evidence stages; the additional README confirmation is reported separately in Section~\ref{sec:readme}. The Data2txt development cohort was already exposed during method development and is not independent confirmation. It also comes from official TRAIN, which the public models may have seen during their original training. The tool-output test measured the fixed \Native{} point before the final shared-engine prefill confirmation. \Prefill{} was not run on that tool-output test. The code-agent selection, readouts, caps, scheduling, metrics, and transferred encoder threshold were fixed before the code-agent examples and labels were opened. We evaluate all selected responses, with no class, quality, length, or outcome filtering.

\begin{table}[t]
\centering\small
\begin{tabular}{lrrr}
\toprule
Source & Responses & Components & Positive\\
\midrule
Data2txt DEV & 1,026 & 171 & 715\\
Tool-output TEST & 617 & 492 & 308\\
Code-agent TEST & 2,015 & 1,613 & 1,014\\
\bottomrule
\end{tabular}
\caption{Evaluation units and released positive labels. DEV is exposed; code-agent is the prospective confirmation. Components are source-derived dependence units, not a proof of independence.}
\label{tab:cohorts}
\end{table}

Both initial test sources are from the pinned public release \path{KRLabsOrg/lettucedetect-code-hallucination}, revision \code{866a7c5392c3cf87e4fb}\allowbreak\code{c2b3808815d524f54331}, under CC BY 4.0~\cite{dataset}. The exact source selectors are \code{lettucedetect-tool-output} and \code{lettucedetect-code-agent}. The authors describe model-proposed local edits with edit-derived reference spans and source-based splits; code-agent TEST uses model-assisted triage and author review/adjudication~\cite[\S4.3--4.4]{grounding}. These are released reference spans, not our independent human-gold annotations. We use them without repairs; they do not establish exhaustive execution or real-world correctness.

For the two initial test sources, components close exact normalized nonempty questions or contexts and context near-duplicates with five-word Jaccard similarity at least 0.85. The development source uses its 171 fixed source families, retaining all six sibling answers per family. Source grouping is completed without labels. Two context-near links from the code-agent selection to the previously exposed tool source are recorded and retained; they are not excluded after selection. These rules do not find all semantic duplicates or establish separation from every model-training example.

\subsection{Models and the fast learned control}

The frozen generative model is the public LettuceDetect Qwen3.5-2B checkpoint, revision \code{4511d95c6a5f24df1e81}\allowbreak\code{0a097685d068208cb149}. The encoder is the released 307M mmBERT-base counterpart, revision \code{8831ce063dff760b01ef}\allowbreak\code{e4519b4188f5ee2456f2}. The paper reports that both are trained on the unified task split~\cite{grounding}. Thus the evaluation is within the authors' held-out task sources, not unknown-domain or contamination-free zero-shot transfer. The models differ in capacity, objectives, and training recipes; we do not attribute their capability difference to early stopping.

The encoder uses native answer-token argmax and flags a response if any valid answer token is unsupported. We also transfer exactly one previously frozen Data2txt threshold: $\max_j(\ell_{j,1}-\ell_{j,0})>-0.9401225447654724$. Native and thresholded predictions share the same forward pass. No threshold is fitted on any test source. On DEV, this fixed threshold is an in-sample diagnostic; an earlier FP32 source-fold threshold control obtained 90.45 F1 and is not replaced by that diagnostic.

The encoder follows the public context-only \code{only\_first} truncation strategy, configured at its 8,192-token capacity rather than the package default 4,096. The full answer IDs and offsets are preserved. Decoder context capacity is 32,768 tokens for both test sources. All overlength encoder contexts remain in the primary evaluation. CPU lengths also prospectively define a secondary subset needing no encoder truncation: 373 tool-output responses in 321 components and 1,574 code-agent responses in 1,189 components. The eligibility rule was fixed before new labels were opened. Actual membership was computed from CPU lengths after local reference projection but before model outputs, without using labels. These subsets do not replace the full cohorts.

\subsection{Execution, uncertainty, and cost}

The shared-engine DEV and code-agent assays initialize the 2B model once. They process original-order chunks of 32, rotating through three arm orders: \Full{}--\Native{}--\Prefill{}, \Native{}--\Prefill{}--\Full{}, and \Prefill{}--\Full{}--\Native{}. The code-agent assay has 189 generate calls and 6,045 decoder requests. Each arm uses greedy decoding, seed 0, the same checkpoint and native source prompt, and unrestricted vocabulary. Prefix caching is explicitly disabled and its actual configuration is recorded. The main confirmation runs on one NVIDIA H100 80GB HBM3 (81,559 MiB reported), driver 610.43.02, with vLLM 0.17.1+cu128, Torch 2.10.0, Transformers 4.57.6, and tokenizers 0.22.2. We use eager vLLM execution with at most eight sequences; the fixed interleaving is not repeated randomized latency sampling. The earlier tool-output arms instead run in separate processes, so their individual loading times are shown separately.

The encoder runs once in a separate process, batch size four, in original order with BF16 parameters and logits, FP32 readback, SDPA, and TF32 disabled. We verify its 22 FP32 rotary buffers and effective rotary theta of 160,000. The existing Transformers 4.57.6 compatibility implementation is held fixed. It is not a claim of bitwise identity with the upstream HF5 implementation or a lower bound on optimized encoder runtime.

Saved token IDs, source hashes, actual samplers, model identities, offsets, request counts, and cost records are checked before evaluator labels are parsed. No incomplete arm receives favorable partial-cohort quality scoring. We retain invalid outputs and report invalid-as-negative, invalid-as-positive, and always-wrong sensitivity rules. All focal reported arms have zero invalid decisions and zero unknown request costs, so these rules give the same scores here.

Our main metric is positive-class response F1; accuracy and confusion counts accompany the artifacts. Paired percentile intervals use 2,000 whole-component bootstrap resamples, seed 20261024, shared across contrasts within a cohort. They condition on the frozen model, grouping, and observed run. They do not account for adaptive DEV use, undetected dependence, or training and timing variation, and are not equivalence tests.

Decoder \emph{service} timers include each chunk's generation and associated local result handling, excluding shared initialization and source preparation. \emph{Engine-call} wall time is the narrower generation-call timer, not a direct GPU-kernel profiler. Encoder forward, tokenization, transfer, projection, initialization, and worker times are reported separately. In particular, its worker-minus-initialization timer includes tokenization and is not interchangeable with decoder service time. Main-job GPU seconds are allocation duration times allocated GPUs, counted once, including startup and idle or CPU work during the allocation.

\section{Detailed frozen measurements}
\label{app:measurements}

\begin{table*}[t]
\centering\small
\begin{tabular}{lrrr}
\toprule
Readout / system & Data2txt DEV & Tool-output TEST & Code-agent TEST\\
\midrule
2B full, official-offset presence & 94.5607 & 90.7285 & 83.4128\\
2B full, raw-list presence & 94.4948 & 90.9091 & 83.5363\\
2B \Native{} & 94.5531 & 90.9091 & 83.6594\\
2B \Prefill{} & 94.3554 & --- & 83.4304\\
307M BF16 encoder, native argmax & 87.3323 & 76.7361 & 76.5936\\
Same encoder, frozen DEV threshold & $90.5292^{\dagger}$ & 78.5603 & 77.3960\\
Always positive & 82.1367 & 66.5946 & 66.9528\\
\bottomrule
\end{tabular}
\caption{Response F1 (percent), all selected responses. Full columns reuse the same generations; encoder native and fixed-threshold rows reuse one forward per response. $\dagger$The fixed threshold was chosen on exposed DEV and is in-sample there, not a new out-of-fold estimate. DEV encoder measurements are cached precision controls, not additional arms of its shared-engine decoder run. A dash means not run.}
\label{tab:quality}
\end{table*}

\subsection{Development and tool-output evidence}

On exposed DEV (Table~\ref{tab:quality}), \Prefill{} is 0.1977 percentage points below \Native{} (95\% interval $[-0.5374,+0.0829]$). There are five branch differences, despite every full and native-eight output sharing the fixed first seven IDs. Service time falls from 34.475 to 14.033 seconds; full service takes 189.105 seconds. Shared initialization is 49.633 seconds and complete decoder worker time is 347.999 seconds. The completed allocation costs 352 GPU-seconds. A preserved sampler-guard failure consumed another 64 GPU-seconds before any model call; its runtime-only repair accepted vLLM's greedy \code{top\_k} normalization without changing the method.

On the tool-output test, all 617 raw full decisions and actual first eight output IDs match \Native{}. Worker time falls from 157.121 to 83.543 seconds, a 46.83\% reduction, with separate loading times of 54.911 and 36.900 seconds. Worker-minus-loading time is 102.210 versus 46.644 seconds; engine-call time is 84.889 versus 29.833 seconds. These are single process observations.

Official location matching removes one true-positive full prediction. Consequently, \Native{} exceeds official full F1 by 0.1806 points, with a component interval of $[0,+0.5733]$, while matching raw full exactly. This difference is a postprocessing effect, not evidence that early stopping reasons better. Its gap over the frozen-threshold encoder is 12.35 points ($[8.93,15.80]$); on the 373-response no-truncation subset, the gap is 9.46 points ($[5.42,13.86]$). The encoder takes 106.125 seconds overall, including 19.366 seconds of tokenization and 69.979 seconds of forward calls. The complete three-system allocation costs 440 GPU-seconds.

\subsection{Prospective code-agent confirmation}

The all-response code-agent results support two distinct efficiency points. \Native{} reaches 83.6594 F1 versus official full's 83.4128. \Prefill{} reaches 83.4304, with precision 90.0571\%, recall 77.7120\%, and accuracy 84.4665\%. Its confusion counts are 788 true positives, 87 false positives, 226 false negatives, and 914 true negatives. The corresponding official-full counts are 787, 86, 227, and 915. Aggregate proximity therefore does not mean identical errors.

\begin{table}[t]
\centering\small
\begin{tabular}{lrr}
\toprule
Code-agent F1 contrast & $\Delta$ (pp) & 95\% interval\\
\midrule
\Prefill{} $-$ full official & +0.018 & $[-0.273,+0.327]$\\
\Prefill{} $-$ \Native{} & $-$0.229 & $[-0.548,+0.059]$\\
\Native{} $-$ full official & +0.247 & $[-0.086,+0.605]$\\
\Prefill{} $-$ encoder fixed & +6.034 & $[+3.969,+8.114]$\\
\Native{} $-$ encoder fixed & +6.263 & $[+4.187,+8.373]$\\
\bottomrule
\end{tabular}
\caption{Paired 1,613-component bootstrap intervals on all 2,015 code-agent responses. The primary prefill comparisons retain possible losses; these are not equivalence intervals.}
\label{tab:ci}
\end{table}

All 2,015 full and \Native{} trajectories have the fixed first seven IDs. Full's first eight IDs match \Native{} on 2,007 responses; reconstructed prefill output also matches \Native{} on 2,007. The eight differences in each comparison show why the ideal same-prefix argument is insufficient to infer empirical quality. None of the prefill continuations uses an unexpected diagnostic branch ID, although the decoder was not restricted to those IDs. Full official processing drops 35 span strings overall and changes two positive response decisions relative to raw presence. We retain the official/raw distinction in every comparison.

The same-training encoder obtains 76.5936 F1 by native argmax and 77.3960 using the transferred threshold. It truncates only the context of 441 responses, removing 1,453,394 encoder tokens; answers remain intact. On the prespecified no-truncation subset of 1,574 responses and 1,189 components, full official, \Native{}, and \Prefill{} obtain 82.3939, 82.6389, and 82.5000 F1, while the fixed-threshold encoder obtains 76.5432. The prefill--encoder gap is 5.9568 points ($[3.6213,8.2334]$). Thus unequal maximum input capacity alone does not explain the observed model gap. This comparison still does not isolate the causal contribution of capacity, training, or architecture.

\subsection{Measured computation and complete costs}

\begin{table*}[t]
\centering\small
\begin{tabular}{lrrrrr}
\toprule
Code-agent decoder readout & Service (s) & Engine calls (s) & Input tokens & Output tokens & Logical total\\
\midrule
Full & 446.407 & 434.534 & 8,605,121 & 95,134 & 8,700,255\\
\Native{} & 99.477 & 88.252 & 8,605,121 & 16,120 & 8,621,241\\
\Prefill{} & 67.668 & 53.419 & 8,619,226 & 2,015 & 8,621,241\\
\bottomrule
\end{tabular}
\caption{One shared resident 2B engine on the complete code-agent source. Service excludes common loading and preparation; engine-call wall time is nested within service. No per-arm fraction of shared initialization or allocated GPU seconds is fabricated. The encoder's separately reported forward timer is not an identically scoped service measurement.}
\label{tab:cost}
\end{table*}

Table~\ref{tab:cost} gives the paired decoder timers. \Native{} reduces service by 77.72\% relative to full generation. \Prefill{} reduces it by 84.84\%, or 31.98\% relative to \Native{}. The narrower engine-call reductions are 79.69\%, 87.71\%, and 39.47\%, respectively. These percentages concern the resident decoder's service, not total job time.

Common decoder initialization takes 74.512 seconds. Source preparation and tokenization take 203.388 seconds, and asset verification takes 3.662 seconds. The full three-arm decoder worker takes 919.893 seconds. Prefill still processes the long evidence. We do not estimate a standalone one-arm deployment time by distributing shared work arbitrarily.

The encoder separately takes 360.969 seconds overall: 11.807 for initialization, 79.818 for tokenization, 234.720 for forward calls, 0.912 for input transfer, and 17.903 for projection and readback, plus remaining verification and bookkeeping. Its worker-minus-initialization time of 349.161 seconds includes tokenization. Comparing that number directly with the decoder's preparation-excluding 67.668-second service timer would exaggerate a common-scope speed claim. The forward/engine-call timers and complete worker totals are exposed instead of collapsed into one purportedly equivalent latency.

The complete shared-decoder-plus-encoder allocation is 1,296 GPU-seconds, counted once, including startup and CPU work during the allocation. Before allocation, decoder and encoder CPU fit checks take 223.27 and 109.92 seconds; the subsequent CPU ID audit takes 231.60 seconds after GPU release. These are additional CPU costs, outside that GPU duration. None adds a model call. The readouts introduce no paid API, new training, or new weights.

The logical total for \Native{} and \Prefill{} is exactly 8,621,241 tokens. Prefill adds 14,105 input tokens and removes exactly 14,105 output tokens. Even relative to full, total logical tokens fall by only 0.91\%, much less than service time. On tool output the total-token reduction is 0.44\%. These quantities demonstrate a change in sequential work, not an equal reduction in a provider's bill. Encoder and decoder tokens use different tokenizers and are not interchangeable monetary units.

\subsection{Prospective developer-document extension}
\label{sec:readme}

We fixed the same readouts, checkpoints, caps, schedule, and encoder threshold before materializing any README instances or labels. All 641 responses of \code{lettucedetect-readme} from the same pinned release remain, grouped into 638 source-derived lexical components. This is a new project-held-out source, within the authors' task: both public models were trained on its TRAIN. The component count does not establish repository or semantic independence.

\begin{table}[t]
\centering\small
\begin{tabular}{lrr}
\toprule
README readout & F1 (\%) & Decoder service (s)\\
\midrule
Full, official/raw & 98.5782 & 85.2763\\
\Native{} & 98.5782 & 22.1382\\
\Prefill{} & 98.5782 & 8.3065\\
Encoder, native & 90.2928 & ---\\
Encoder, frozen threshold & 89.0855 & ---\\
\bottomrule
\end{tabular}
\caption{Complete README source, with no invalid outputs or unknown request costs. Decoder rows share one resident engine; the two encoder rows reuse the same forward calls. Encoder timers have a different scope and are reported separately in the text.}
\label{tab:readme}
\end{table}

All 641 official/raw full decisions, \Native{} decisions, and \Prefill{} decisions agree. Every native prefix and prefill reconstruction also agrees; these are sample observations, not population equivalence. \Prefill{} reduces decoder service by 90.2593\% against full and 62.4789\% against \Native{} (Table~\ref{tab:readme}). All encoder pairs fit without truncation. Its lower F1 therefore cannot be attributed solely to the 8k limit, but the model capability gap remains inherited. The transferred threshold performs worse than native argmax and is retained without refitting.

Shared decoder initialization takes 57.5152 seconds, tokenization/preparation 17.1854, and the complete three-arm worker 207.5509. The encoder's complete worker takes 19.0449 seconds, including 8.7356 for initialization and 2.0495 for tokenization; its forward calls take 3.1505 seconds. In particular, its 10.3093 seconds excluding initialization still includes preparation and cannot be compared directly with \Prefill{}'s preparation-excluding 8.3065-second service as a speed win. The combined allocation costs 232 GPU-seconds, counted once; post-release CPU actual-ID auditing takes another 25.697 seconds. \Native{} and \Prefill{} each use 502,567 logical tokens, versus 516,274 for full: only a 2.655\% total-token reduction, without a proportional billing claim.

The released card describes README as lower-precision first-pass edit-derived references, with some errors detectable without context~\cite{dataset}. The code-test per-example review description does not extend to this source. Thus 98.58 F1 is agreement with released references, not independently established real-error accuracy, evidence dependence, contamination-free transfer, or field-wide state of the art. This source is now closed to method tuning.

\subsection{An optional first predicted location}

On exposed DEV, the fixed \FirstQuote{} rule gives alarm F1 94.3554 versus matched full raw 94.4212, with service 69.070 versus 194.379 seconds ($-64.47\%$). We then transfer that rule unchanged to all 2,015 code-agent responses in 1,613 components. Their earlier outcomes are already known: this is an additional functional comparison, not new independent confirmation or a replacement for Tables~\ref{tab:quality}--\ref{tab:cost}.

\begin{table}[t]
\centering\small
\begin{tabular}{lrrr}
\toprule
Same-$s$ readout & Alarm F1 & Service (s) & Located/alarms\\
\midrule
Full & 83.5363 & 425.7906 & 874/875\\
\FirstQuote{} & 83.5363 & 204.6371 & 861/875\\
\Prefill{} & 83.4304 & 63.9525 & ---\\
\bottomrule
\end{tabular}
\caption{Additional matched-prefix, matched-detokenization code-agent run. Full returns its first valid official offset; \FirstQuote{} returns an optional first-quote offset, and \Prefill{} is alarm-only. All 2,015 responses remain.}
\label{tab:firstquote}
\end{table}

Full raw and \FirstQuote{} agree on every response alarm in this run (Table~\ref{tab:firstquote}). \FirstQuote{} reduces service by 51.9395\% and output tokens from 81,036 to 32,606 ($-59.7636\%$), but total tokens by only 0.55665\%; each arm processes 8,619,226 input tokens. This observed agreement does not establish population equivalence.

Location coverage declines: 14 of 875 \FirstQuote{} alarms are unmappable, versus one for full. Among 861 jointly located predictions, 846 quotations and offsets agree and 15 differ; 13 locations are available only after full generation. Offset-presence F1 is 82.8800 versus full official 83.4746, a $-0.5946$-point difference (component interval $[-1.0496,-0.2048]$). Against released edit-derived spans, first-location micro character precision is 56.6728\% versus 56.8585\%, and recall over all reference characters is 28.6444\% versus 29.1267\%. A located quote is a predicted error location, not an entailment certificate or a complete span set.

All 6,045 requests pass the actual-ID/hash audit with zero invalid alarms or unknown costs. Shared initialization is 56.8347 seconds, preparation 228.7269, asset verification 5.3583, and worker time 1,008.4328. The complete allocation costs 1,017 GPU-seconds, counted once; the post-release CPU audit takes 188.2404 seconds. These new matched-run clocks do not replace the prospective main experiment's costs.

\subsection{Bounded warm single-request measurements}

We hash-select 18 already exposed Data2txt DEV components and one answer each, without labels. Three rounds per case--arm pair balance all three order positions; a nineteenth component provides one warmup per arm. All 165 calls remain: 162 measured plus three warmup. One resident engine uses maximum sequence count one and no prefix caching. Base input lengths are 938--1,921 tokens (median 1,362); prefill adds seven.

\begin{table}[t]
\centering\small
\begin{tabular}{lrr}
\toprule
Readout & Host generate & Service\\
 & median / p95 (ms) & median / p95 (ms)\\
\midrule
Full & 797.593 / 4,401.658 & 813.226 / 4,419.220\\
\Native{} & 175.062 / 190.273 & 192.419 / 212.430\\
\Prefill{} & 29.634 / 33.179 & 46.923 / 54.282\\
\bottomrule
\end{tabular}
\caption{Descriptive warm single-request clocks on old DEV: 54 calls per arm, but only 18 source units, each repeated three times. Three warmup calls are excluded here and charged separately. These p95 values are not production tail-latency estimates.}
\label{tab:single}
\end{table}

For each case we take each arm's median over three rounds, divide prefill by native-eight latency, then summarize those 18 ratios by their median, rather than dividing Table~\ref{tab:single}'s pooled medians. The host-generate ratio is 0.170154 (82.9846\% reduction), median paired difference $-145.517$\,ms; service gives 0.246363 (75.3637\%), difference $-145.485$\,ms. All 54 round-wise triples have valid, agreeing decisions and identical reconstructed eight-token prefixes. This is functional agreement, not new F1 or quality-equivalence evidence.

Host generation times synchronous \code{llm.generate}. Service adds actual re-encoding, prompt writing, optional existing-driver-context synchronization, parsing, and completion/prediction writes; it includes the pre-generate inflight-session write but excludes final request/batch/session record commits and does not independently synchronize another process. Shared initialization takes 50.699 seconds, preparation 18.812, verification 4.005, and the complete worker 183.567. Warmup service totals 4.205 seconds; all warmup is charged in 189 GPU-seconds. Subsequent CPU ID auditing takes 19.866 seconds outside that allocation. Eighteen sources, not 54 independent cases, support only this bounded host-clock observation, not production tails, general modern interactive latency, or end-to-end agent latency.

\paragraph{Modern code-agent extension.}
We repeat the same finite singleton design on the existing code-agent cohort, with a new fixed hash salt selecting 18 source components, one answer per component, and a nineteenth warmup component. Selection uses only component keys and qids, with no labels, input-length selection, or method changes. The source's historical quality results are already known; this is an additional timing measurement, not new quality confirmation. We retain all 165 calls and use the original code-agent user wrapper, one resident H100 80GB, maximum sequence count one, 32k context, BF16 eager execution, and no prefix caching. Selected base prompts span 2,478--7,194 tokens (median 4,203), extending beyond the short DEV input range.

\begin{table}[t]
\centering\small
\begin{tabular}{lrr}
\toprule
Readout & Host generate & Service\\
 & median / p95 (ms) & median / p95 (ms)\\
\midrule
Full & 281.973 / 3,141.986 & 347.619 / 3,212.313\\
\Native{} & 202.756 / 208.772 & 254.792 / 286.389\\
\Prefill{} & 36.912 / 52.023 & 86.224 / 127.766\\
\bottomrule
\end{tabular}
\caption{Warm singleton clocks on 18 hash-selected code-agent source components, each repeated three times. All 54 measured calls per arm remain; three separately charged warmup calls are excluded from these summaries. Descriptive p95 is not a production tail estimate.}
\label{tab:single_code_agent}
\end{table}

The prespecified median of 18 per-qid median-of-three prefill/native-eight ratios is 0.176369 for host generation (82.3631\% reduction), with median paired difference $-166.345$\,ms. Service gives 0.333779 (66.6221\%), difference $-164.449$\,ms. These are paired summaries, distinct from dividing Table~\ref{tab:single_code_agent}'s pooled medians. All 54 measured triples have valid, agreeing branch decisions, and all actual full-first-eight/native-eight/prefill-reconstructed IDs agree. This functional observation does not supersede the eight decision changes in the complete 2,015-response batch confirmation or establish equal population quality.

Shared initialization takes 65.283 seconds, tokenization/preparation 24.704 seconds, asset verification 6.888 seconds, and complete worker time 219.625 seconds. All three warmup calls cost 3.970 seconds of service and remain charged in the 225 GPU-second allocation. The subsequent CPU actual-ID audit takes 20.893 seconds after GPU release. Measured native-eight and prefill logical totals are both 240,888 tokens; 378 output positions become input positions. The bounded gain extends to these longer modern inputs; it does not establish cold-start, network, concurrent-service, or complete-agent latency savings.

\paragraph{SDPA encoder singleton control.}
We then run the fixed 307M BF16 encoder on exactly these 18 sources and rounds, with the same nineteenth-source warmup: 55 forwards in a separate allocation, without new selection, labels, F1, or fitting. Each binary service call freshly encodes the pair, constructs and transfers tensors, synchronizes the forward, reduces FP32 answer-logit margins on GPU, reads one scalar, and commits a minimal alarm. Native and transferred-threshold decisions share that forward. Full logits, actual IDs/masks, and exact CPU readout checks are recorded outside binary service, with their costs retained.

Encoder median service is 273.088\,ms (forward 240.680\,ms; encoding 29.073\,ms). The prespecified median of 18 per-qid median-of-three encoder/decoder service ratios is 0.536918 against full, 1.080135 against \Native{}, and 3.205585 against \Prefill{}; the latter median paired difference is $+189.084$\,ms. All 55 scalar/full-logit audits agree and all three-round encoder decisions are stable. This locates a useful warm operating point, not new quality confirmation. Four of the 18 encoder contexts are truncated at 8,192 tokens with complete answers retained; tokenizers, backends, allocations, preprocessing, and logging boundaries differ, and historical decoder service includes heavier logging. Encoder loading is shorter: initialization 14.952\,s, preparation 0.806\,s, assets 1.520\,s, warmup service 0.809\,s, and worker 35.201\,s. Measured diagnostic work adds 0.372\,s outside binary service; the whole allocation is 37 GPU-seconds, counted once. These clocks do not establish cold-start or universal dominance.

\paragraph{Attention implementation control.}
We reuse the same 18 sources, three rounds, and nineteenth-source warmup in one resident encoder, rotating three fixed implementations (Table~\ref{tab:encoder_attention}). All retain the checkpoint, BF16, original QKV and FP32 RoPE, token IDs, answer mask, and 8k context policy. Two project bridges use the installed FA3 kernel, not native HF FlashAttention2: one retains dense-mask construction; the other expresses the same global/local windows directly and skips redundant masks only for these batch-one, all-valid inputs.

\begin{table}[t]
\centering\small
\begin{tabular}{lrr}
\toprule
Encoder implementation & Service (ms) & Forward (ms)\\
\midrule
Stock SDPA & 276.547 & 238.149\\
FA3, dense masks retained & 233.254 & 203.850\\
FA3, direct windows & 44.121 & 13.248\\
\bottomrule
\end{tabular}
\caption{Warm call medians on the fixed 18-source plan: 54 measured forwards per implementation, plus one charged warmup each. Service includes fresh encoding, transfer, forward, binary readout, and minimal recording. No labels or new F1 are used.}
\label{tab:encoder_attention}
\end{table}

The prespecified median paired direct-window/SDPA service ratio is 0.171300. Against historical \Prefill{} it is 0.497155, reversing the SDPA speed ordering, subject to the cross-allocation and service-scope differences above. All 54 paired native and frozen-threshold decisions agree across implementations, but margins change on all 18 sources; the two FA3 variants agree with each other. Fixed zero-QK primitives validate window support and normalization, not arbitrary-QK or quality equivalence. Dense-mask host clocks are nested enqueue timings, not additive GPU-kernel profiles. The new allocation costs 70 GPU-seconds once, including diagnostics; initialization takes 28.503\,s, preparation 0.998\,s, and complete worker time 67.683\,s. This bounded timing result does not by itself establish the bridge's complete-cohort quality.

\paragraph{Complete optimized encoder control.}
A separately frozen pass now evaluates the direct-window bridge on all original 2,015 code-agent responses and 1,613 components, in original order, with no labels, fitting, new cases, or decoder replay. Both fixed readouts share each batch-one forward. The same 441 context truncations and all answer IDs/offsets remain. This adds a stronger implementation control on an already exposed test, not independent confirmation.

\begin{table}[t]
\centering\small
\begin{tabular}{lrr}
\toprule
Encoder implementation & Native & Fixed\\
\midrule
BF16 SDPA (batch 4) & 76.5936 & 77.3960\\
BF16 FA3/windows (batch 1) & 76.7976 & 77.3345\\
\bottomrule
\end{tabular}
\caption{Complete code-agent response F1 (percent). Native and the sole old-DEV threshold share each implementation's forward; neither is refitted. Earlier decoder results remain unchanged.}
\label{tab:encoder_fa3_full}
\end{table}

All 2,015 actual-input, full-logit, scalar, and window-execution audits pass, with no invalid or unknown-cost calls. Relative to cached SDPA, 1,952 margins and 16 native/12 fixed-threshold decisions change; native F1 rises 0.2040 points while fixed-threshold F1 falls 0.0615 points (Table~\ref{tab:encoder_fa3_full}). Batch shape and attention arithmetic both differ, so these changes are not attributable solely to the kernel and do not establish quality equivalence. \Prefill{} exceeds the optimized fixed-threshold encoder by 6.0958 points (1,613-component 95\% interval $[4.0262,8.2156]$); on the unchanged 1,574-response no-truncation subset, it obtains 82.5000 versus 76.4602, a 6.0398-point gap ($[3.6800,8.3340]$). The fixed-threshold encoder has higher recall (84.12\% versus 77.71\%); the F1 advantage is not every-metric superiority and remains inherited from the public models.

The new encoder service totals 100.167\,s, including 65.800\,s of fresh pair encoding and 28.406\,s of synchronized forward calls; complete logits/ID diagnostics add a separately measured 15.441\,s. Initialization is 21.400\,s, assets 2.477\,s, and worker time 169.505\,s; the complete allocation is 173 GPU-seconds once. Post-release local scoring takes 17.104\,s without model calls. These fresh-encoding service totals and the decoder's preparation-excluding totals have different scopes and cannot be divided into a common-protocol end-to-end speedup. The same-source singleton result instead shows a faster optimized encoder point with lower observed quality than the 2B readout, while the within-2B decoding increment remains.

\subsection{Direct one-token binary control on exposed DEV}

A shorter direct binary answer is an important simple alternative to native structured readout. We prospectively fix one control on all 1,026 already exposed Data2txt DEV responses (171 source components), using the same frozen 2B checkpoint, facts, user wrapper, and tokenizer. Its shorter system instruction asks for Yes if any answer content is unsupported and No otherwise. Only this control restricts generation to the two exact single-token verbalizers, Yes (ID 9175) and No (ID 2665), with temperature-zero argmax and no fitted threshold. Both actual assistant continuations are CPU-qualified on every input before model construction. The native readouts remain unmasked. No closed test is used to select or optimize the control.

\begin{table}[t]
\centering\small
\begin{tabular}{lrrr}
\toprule
DEV readout & F1 & Service (s) & Engine calls (s)\\
\midrule
\Native{} & 94.487 & 44.433 & 35.926\\
\Prefill{} & 94.355 & 17.098 & 10.048\\
Direct binary1 & 3.026 & 16.095 & 9.530\\
\bottomrule
\end{tabular}
\caption{Additional fixed DEV control in one new resident engine. All 1,026 responses per arm remain, with zero invalid decisions. These new observations do not replace the earlier DEV comparison or add independent quality confirmation.}
\label{tab:binary}
\end{table}

All 3,078 requests and 99 rotating batch calls complete. Binary1 predicts only 12 positive responses: 11 true positives and one false positive, with 704 false negatives and recall 1.54\%. Its shorter input costs 1,131,895 tokens, versus 1,471,501 for \Prefill{}; both generate 1,026 tokens. It saves 1.004 seconds of service relative to \Prefill{} at unusable detection quality (Table~\ref{tab:binary}). Thus enforcing a valid answer format does not make this direct-prompt transfer task-competent. This fixed recipe is closed; it does not show that binary classification or every binary prompt is weak.

Within this run, \Prefill{} reduces service by 61.52\% against \Native{}, with F1 lower by 0.1317 points (171-component interval $[-0.4156,+0.1436]$). All native outputs share the fixed first seven IDs; four reconstructed prefill trajectories differ. Shared initialization takes 66.884 seconds, tokenizer/preparation clock 56.772 seconds (including 23.393 seconds for tokenizer initialization), and asset verification 4.075 seconds. Per-arm preparation takes 10.860/11.546/10.972 seconds; these nested clocks are not added again. The complete worker takes 230.005 seconds and allocation 236 GPU-seconds; subsequent CPU ID auditing takes 48.223 seconds after GPU release. Native8 remains the strongest useful same-model short-output control. This format-sensitive result supports reusing the learned native branch, without isolating a causal role for localization training.

\subsection{Observed length dependence}

A posthoc, zero-model-call analysis retains all 63 paired code-agent chunks, including the final 31-response chunk. The prefill/native-eight engine-call ratio correlates with mean prompt length (tie-aware Spearman 0.923; service 0.881). Four fixed length-rank bins have median generation ratios 0.5665, 0.5969, 0.6389, and 0.6673: relative savings decrease with length. Generation is faster in all 63 chunks; service in 62, retaining one ratio-2.411 outlier. Arm-position summaries vary much less. This is consistent with prompt work amortizing saved serial decode steps, but it remains observational rather than causal or kernel-level evidence; it does not explain cross-workload differences by length alone.

\section{Auxiliary encoder-to-generator routing}
\label{app:cascade}

This extension was explored after the structural-readout study. It reuses the frozen models and is not a new routing algorithm: confidence and verification cascades have established precedents \citep{frugalgpt,automix}. We retain it as an auxiliary detector-specific allocation point rather than the premise of the main contribution.

\paragraph{Frozen gate and strong control.}
On the exposed Data2txt development set, a response escalates to \Prefill{} when the absolute native encoder margin is no larger than a fixed cutoff. Five whole-source hash folds retain cutoffs selected from the old SDPA training folds: 1.2890625, 1.205078125, 1.236328125, 1.17578125, and 1.244140625. The subsequent FA3 implementation does not refit them. Unescalated predictions use the sole old-DEV threshold $-0.9401225447654724$. The control uses the same calibrated encoder and exactly the same escalation count in each fold, choosing the smallest fixed-salt identifier hashes. All contrasts preserve all responses; there is no budget, salt, seed, or target-threshold search. The calibrated hash is the relevant strong control, rather than a weaker native-zero fallback.

\paragraph{Actual development services.}
One allocation keeps both models resident on one GPU, with a fixed 0.70 vLLM memory share, one active decoder sequence, BF16 precision, and no prefix cache. A separately charged label-free encoder pass freezes actual per-fold counts and hash routes before any new decoder output. Three policies then alternate in original row order. Both cascade policies pay fresh encoder tokenization, transfer, forward, FP32 margin projection, and routing on every response, including non-escalated responses. Only escalated cases additionally pay fresh decoder tokenization/generation/parsing. Always-prefill pays only its own required stages. All policies stop their service clocks after the same minimal alarm write and fsync. Full-logit/ID diagnostics, source qualification, initialization, probe, warmup, and cleanup are charged separately; they are not declared saved by the route.

\begin{table}[t]
\centering\small
\begin{tabular}{lrrr}
\toprule
DEV policy & F1 & 2B calls & Service (s)\\
\midrule
Always \Prefill{} & 94.4212 & 1,026 & 48.3604\\
Uncertainty route & 92.5395 & 252 & 30.8888\\
Calibrated hash & 91.7942 & 252 & 30.8996\\
\bottomrule
\end{tabular}
\caption{Actual co-resident services on exposed DEV, 1,026 responses/171 components. This is an exploratory quality--service tradeoff, not independent confirmation.}
\label{tab:cascade_dev}
\end{table}

Uncertainty exceeds the calibrated equal-call hash by 0.7453 F1 points (conditional component interval $[-0.2544,+1.8748]$), while its service total is nearly the same. Against always-\Prefill{}, it loses 1.8817 F1 points ($[-2.8047,-0.9790]$) and reduces aggregate service by 36.1279\% (ratio 0.638721, interval $[0.607873,0.670705]$). Its descriptive p95 service is worse, 67.047 versus 53.677\,ms, despite a lower median. Uncertainty yields 645/34/70/277 TP/FP/FN/TN, versus 660/63/55/248 for hash: precision improves but recall falls. Neither equal quality nor tail-latency dominance follows.

Both routed policies make 252 actual calls (24.5614\%), with no route drift, invalid decisions, unknown costs, or audit issues. Relative to the old SDPA cache, 962 encoder margins and eight native/four calibrated decisions change; actual always-\Prefill{} differs from its old cache on three decisions. The actual counts and quality therefore do not substitute cached values. All 1,531 actual decoder calls pass the post-release CPU ID audit. The full allocation is 327 GPU-seconds, including both initializations (58.4077+2.2011\,s), preflight (12.9608\,s), assets (5.6857\,s), probe service (21.9523\,s), warmups, diagnostics, planning, primitives, and cleanup; worker time is 324.3946\,s. Fewer decoder calls do not imply proportional billed GPU savings.

An initial local validator used encoder-style compact JSON hashing for the decoder's native prompt-ID serialization and refused all calls before parsing labels. A separately frozen scorer corrects only that serialization expression. The original source, refusal, repair record, and raw outputs remain; no method, mask, metric, data, or model call was changed.

\paragraph{Fixed modern transfer replay.}
The entire gate is transferred once to the already evaluated code-agent cohort using its qualified FA3 margins and original \Prefill{} decisions. No code-agent quantile is fitted. This is a CPU combination of existing outputs, not actual cascade execution or new independent confirmation.

\begin{table}[t]
\centering\small
\begin{tabular}{lrr}
\toprule
Code-agent cached policy & F1 & Planned 2B calls\\
\midrule
Always calibrated encoder & 77.3345 & 0\\
Always \Prefill{} & 83.4304 & 2,015\\
Uncertainty route & 82.0459 & 596\\
Calibrated hash & 79.2220 & 596\\
\bottomrule
\end{tabular}
\caption{Fixed transfer replay on all 2,015 responses/1,613 components. Intended calls are not measured modern service savings. Original context truncations and complete answers remain.}
\label{tab:cascade_transfer}
\end{table}

The route exceeds equal-call hash by 2.8239 F1 points ($[1.3801,4.3363]$), calibrated encoder by 4.7114 ($[3.1638,6.2716]$), and falls below always-\Prefill{} by 1.3845 ($[-2.8190,+0.0376]$). All 596 planned escalations (29.5782\%) use unchanged cutoffs and salts. Uncertainty and hash share 194 escalations and each has 402 unique ones. Their TP/FP/FN/TN are 786/116/228/885 and 835/259/179/742. As on DEV, precision and recall trade off. This replay takes 3.3197 CPU seconds, with zero new model calls, GPU seconds, or API expenditure.

\paragraph{Actual modern cascade services.}
A subsequently completed fixed assay executes all three policies on the same 2,015 code-agent responses and 1,613 components (Table~\ref{tab:cascade_code_agent}). It retains the DEV gate, threshold, salts, all 441 encoder context truncations, and complete answers; no new code-agent threshold or budget is fitted. A label-free FA3 probe fixes the actual per-fold counts and hash routes before any new decoder output. Both models remain resident under the same fixed memory share as the DEV assay, and policies rotate once per response in original order. Both routed policies pay fresh encoder processing and routing on every response, with fresh decoder processing only when escalated; always-\Prefill{} pays its own required stages. The shared service boundary includes encoding, transfer, required forwards, readout, routing, parsing, and minimal alarm writing/fsync. This is actual execution on a previously exposed cohort, not new independent confirmation.

\begin{table}[t]
\centering\small
\begin{tabular}{lrrr}
\toprule
Code-agent policy & F1 & 2B calls & Service (s)\\
\midrule
Always \Prefill{} & 83.4921 & 2,015 & 150.5046\\
Uncertainty route & 82.0459 & 596 & 140.9961\\
Calibrated hash & 79.2220 & 596 & 140.8218\\
\bottomrule
\end{tabular}
\caption{Complete actual modern cascade services, with no invalid decisions, unknown costs, or route drift. Both cascades use 596 calls (29.5782\%); their encoder and routing costs are included. These separately measured clocks do not replace the prospective readout study.}
\label{tab:cascade_code_agent}
\end{table}

Uncertainty exceeds calibrated equal-call hash by 2.8239 F1 points (1,613-component conditional 95\% interval $[1.3801,4.3363]$) at almost identical service: the ratio is 1.001237 ($[0.969791,1.031352]$), so the route is not faster than hash. Against actual always-\Prefill{}, uncertainty loses 1.4461 F1 points ($[-2.8776,-0.0135]$), while reducing aggregate service by 6.3178\% (ratio 0.936822, $[0.912262,0.959927]$). This is a smaller measured saving than DEV's 36.1279\%, despite 70.4218\% fewer decoder calls. Fresh encoder work and the escalated decoder work both count. Uncertainty precision/recall are 87.1397/77.5148\%, versus hash's 76.3254/82.3471\%; the precision gain retains a recall loss. Descriptive service p95 is worse than always-\Prefill{}, 165.158 versus 125.966\,ms. Intervals use 2,000 shared component resamples (seed 20261025) and condition on this one host/pass, actual routes, and frozen outputs; they do not estimate repeated scheduling uncertainty or production tails.

The probe exactly preserves the cached FA3 margins and both encoder readouts. Actual routes match all planned routes. Uncertainty and hash retain their replay confusion counts; actual always-\Prefill{} changes one cached decision, adding one true positive (789/87/225/914 TP/FP/FN/TN). Its new 83.4921 F1 is therefore reported with the new clock rather than substituting the main experiment's 83.4304 F1. All 8,062 requests complete, including 2,015 probe requests, 6,045 policy requests, and two warmups, with 6,046 encoder forwards. All 3,208 decoder calls, including warmup, pass the post-release CPU actual-ID audit in 88.452\,s; frozen local scoring adds 21.991 CPU seconds without model calls.

The complete allocation costs 999 GPU-seconds once (worker 994.4089\,s), including decoder/encoder initialization (60.7167/2.2613\,s), tokenization preflight (67.7822\,s), assets (5.7789\,s), label-free probe service/diagnostics (98.4331/12.9180\,s), encoder/decoder warmup service (0.3526/3.2833\,s), full-ID/logit diagnostics for the three policies (4.7253/13.5760/14.1633\,s), planning (0.1955\,s), primitive qualification (0.1687\,s), and cleanup (1.1380\,s). These phases are retained separately rather than counted as warm deployment savings; nested clocks are not added again. With both models resident in one allocation, we do not manufacture per-policy GPU bills or infer billed savings from call counts. The useful increment is better allocation of the existing detector's capability than equal-call hash, with a measured quality--service tradeoff against always-\Prefill{}. This auxiliary result does not alter the native-readout contribution or establish quality equivalence or field-wide state of the art.

\CenteredCascadeDevelopmentVnine

\begin{figure}[t]
\centering
\includegraphics[width=\linewidth]{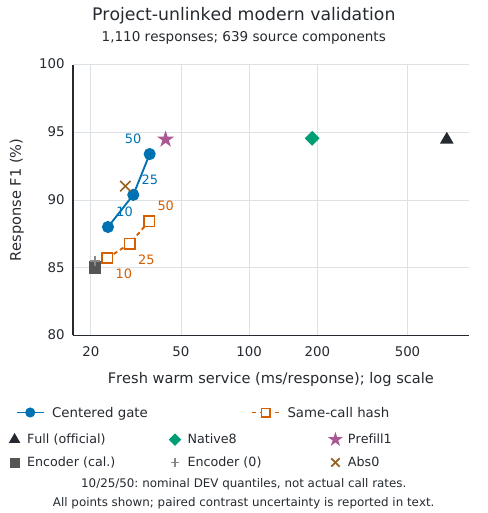}
\caption{All prespecified modern quality--service points, including both
encoder thresholds and every DEV routing radius. Lines connect fixed budgets;
no validation-selected optimum is fitted. These are aggregate estimates;
paired contrast intervals accompany the complete tables.}
\label{fig:cascade-vnine}
\end{figure}

\subsection{Prospective modern confirmation of the complete policy matrix}
\label{app:modern-centered-confirmation}
\label{app:modern-confirmation-vnine}

\paragraph{Selection and reference boundary.}
Before the new source strings or labels were materialized, selection fixed
all 1,117 responses from the code-agent, developer-tool-output, and README
sources of the same pinned release's validation split. Source-only closure
joined normalized exact question/context matches, context near-duplicates
with five-word Jaccard similarity at least 0.85, and whitelisted
instance/document identities. It
also linked against the already exposed DEV and three historical modern TEST
cohorts, reusing their components. Seven tool responses in five components
link to the earlier tool cohort; all remain in execution and the complete
secondary evaluation. The fixed project-unlinked primary contains 1,110
responses in 639 components. No reference, prediction, length, or observed
outcome determined these populations.

Table~\ref{tab:modern-cohorts-vnine} retains all source counts. These units
are a conservative lexical/instance/document closure, not proof of semantic
or repository independence. Tool records have no qualified identity field
under the fixed projection. Upstream mmBERT checkpoint selection explicitly
used development hallucinated-token F1 \citep[Section~6.1]{grounding}; the encoder
comparison therefore reuses upstream development. The generator's analogous
selection rule is unspecified. This is project-unopened validation rather
than a contamination-free external test. Its machine-generated references do not
inherit the release's individual review claim for code-agent TEST; README
remains lower-precision edit-derived evidence. All labels are the original
released span lists, with exact integer bounds checked. Their presence is the
response target, without a new factual-truth or entailment adjudication.

\begin{table*}[t]
\centering\small
\begin{tabular}{lrrrrrr}
\toprule
Source & All $n$ & All $G$ & All positive & Primary $n$ & Primary $G$ & Primary positive\\
\midrule
Code-agent & 190 & 176 & 95 & 190 & 176 & 95\\
Tool output & 240 & 222 & 120 & 233 & 217 & 116\\
README & 687 & 246 & 343 & 687 & 246 & 343\\
Total & 1,117 & 644 & 558 & 1,110 & 639 & 554\\
\bottomrule
\end{tabular}
\caption{Frozen source-only populations. The seven linked tool responses are retained in the complete secondary population; all sources are reported without outcome-based replacement.}
\label{tab:modern-cohorts-vnine}
\end{table*}

\paragraph{Frozen deployment and common service.}
The complete eleven-arm assay uses the fixed DEV threshold
$-0.935546875$, centered radii $0.541015625/1.322265625/2.384765625$
for the prespecified 10/25/50\% DEV quantiles, and uncentered radius
$1.2578125$. No modern quantile or threshold is fitted. Each hash route
matches its centered route's decoder count within source, existing component
hash-fold, and exposure-status strata. The exposure stratum makes matching
hold in both the primary and complete populations. The primary operating
point remains Centered25 even when a sensitivity point looks better.

One H100 allocation keeps the unchanged 2B decoder and optimized BF16 FA3
encoder resident. All eleven arms freshly prepare each required model through
the same helper, with one active decoder sequence, 32k decoder capacity, no
prefix cache, seed zero, and unrestricted greedy decoding. Full/native-eight/
prefill caps are 1,024/8/1. The encoder retains its 8k context-only truncation
and complete answer offsets. A separately paid full encoder probe makes the
label-free hash plan durable before decoder output; the eleven policy orders
then rotate once per original-order response. Fresh services include required
encoding, transfer, forward/generation, readout, routing, parsing, and minimal
alarm write/fsync. Warmup, initialization, full qualification, probe, detailed
diagnostics, final flushing, and cleanup retain separate clocks and remain
paid in the allocation.

\paragraph{Complete quality and native readout.}
Table~\ref{tab:modern-quality-vnine} reports precision, recall, and response
F1 for every native view and all three routed budgets, on both populations.
The calibrated fallback is the same for Abs0, centered, and hash arms. Native
zero is also reported from the unchanged encoder forward; on the primary its
85.5072 F1 is higher than the transferred DEV-threshold encoder's 85.0168.
Thus the calibrated fallback does not dominate even this simple alternative
on the new distribution.

\begin{table*}[t]
\centering\small
\begin{tabular}{lrrrrrr}
\toprule
Readout/policy & Primary $P$ & Primary $R$ & Primary F1 & All $P$ & All $R$ & All F1\\
\midrule
Full, official & 96.7803 & 92.2383 & 94.4547 & 96.7925 & 91.9355 & 94.3015\\
Full, raw & 96.7864 & 92.4188 & 94.5522 & 96.7985 & 92.1147 & 94.3985\\
\Native{} & 96.7864 & 92.4188 & 94.5522 & 96.7985 & 92.1147 & 94.3985\\
\Prefill{} & 96.4286 & 92.5993 & 94.4751 & 96.4419 & 92.2939 & 94.3223\\
Encoder, DEV threshold & 79.6530 & 91.1552 & 85.0168 & 79.7488 & 91.0394 & 85.0209\\
Encoder, native zero & 85.8182 & 85.1986 & 85.5072 & 85.8951 & 85.1254 & 85.5086\\
Abs0 & 93.3586 & 88.8087 & 91.0268 & 93.3837 & 88.5305 & 90.8924\\
Centered10 & 84.2975 & 92.0578 & 88.0069 & 84.3750 & 91.9355 & 87.9931\\
Centered25 & 88.4283 & 92.4188 & 90.3795 & 88.4880 & 92.2939 & 90.3509\\
Centered50 & 93.6479 & 93.1408 & 93.3937 & 93.6709 & 92.8315 & 93.2493\\
Hash10 & 81.0289 & 90.9747 & 85.7143 & 81.1200 & 90.8602 & 85.7143\\
Hash25 & 83.2504 & 90.6137 & 86.7761 & 83.3333 & 90.5018 & 86.7698\\
Hash50 & 86.5285 & 90.4332 & 88.4378 & 86.5979 & 90.3226 & 88.4211\\
\bottomrule
\end{tabular}
\caption{All planned response-quality points (percent). Full official/raw share one generated output. The two encoder thresholds share one forward. These evaluation aliases add no physical calls. Primary and complete populations overlap and are not separate executions.}
\label{tab:modern-quality-vnine}
\end{table*}

On the primary, \Prefill{} has 94.4751 F1 versus full official 94.4547
and native-eight/raw full 94.5522. Its differences are $+0.0204$ points
against full official and $-0.0770$ against native-eight. Shared complete
fresh service falls 827.4728 to 47.4990\,s against full (94.2597\%) and
210.7107 to 47.4990\,s against native-eight (77.4577\%). Table~\ref{tab:modern-contrasts-vnine}
gives conditional component intervals, rather than an equivalence claim.
The primary \Prefill{} confusion counts are 513/19/41/537 TP/FP/FN/TN;
full official has 511/17/43/539. Similar aggregate F1 therefore retains
changed errors.

Across all 1,117 responses, full raw and native-eight decisions agree
exactly, and all native-eight outputs retain the fixed seven syntax IDs.
Prefill changes three native-eight decisions. Official quote-to-offset
mapping changes one full raw decision; the complete full run maps 820
quotation spans and drops seven unmappable spans, with zero unparseable
full outputs. These are parser/trajectory observations, not proof of
entailment. The eight-token and one-token cap hits are intended readout
limits; no full generation reaches its 1,024-token cap. There are zero
invalid decisions, unknown request costs, or route drifts in the completed
matrix, so the three invalid-handling sensitivity rules agree.

\paragraph{Primary routing result and the negative transfer.}
Centered25 obtains 90.3795 F1 with 250 calls, versus exact-call Hash25's
86.7761: $+3.6034$ points, interval $[2.1127,5.2207]$. It also increases
service 32.9830 to 34.1989\,s (ratio 1.036865). Against always-\Prefill{},
it loses 4.0956 F1 points while reducing warm service 28.0009\%
(ratio 0.719991). Against calibrated encoder it gains 5.3627 points while
using 47.7875\% more service. All of these costs include the fresh encoder
on every routed-policy response.

The important counterexample is the simpler Abs0 control. It reaches
91.0268 F1 with only 216 calls and 31.5589\,s. Centered25 is 0.6473 F1
points lower and 8.3651\% slower, while making 34 more calls. Centering
therefore does not transfer as an F1--service improvement over this strong
control, despite its positive DEV result and advantage over random hash
allocation. Its precision/recall of 88.4283/92.4188\% differs from Abs0's
93.3586/88.8087\%: a recall increase carries a substantial precision cost.
This negative transfer is retained rather than repaired by choosing another
threshold, source, or budget.

Centered10 and Centered50 remain prespecified curve points, with
88.0069/93.3937 F1, 102/413 calls, and 26.4375/40.3954\,s respectively.
Both exceed their equal-call hashes, but the 50\% point does not replace
the primary 25\% point after outcomes. Centered50 still trails
always-\Prefill{} in pooled F1. Table~\ref{tab:modern-cost-vnine} gives
actual costs and tails for the complete matrix.

\begin{table*}[t]
\centering\small
\begin{tabular}{lrrrrrr}
\toprule
Physical policy & \shortstack{2B calls\\primary/all} & \shortstack{Encoder input\\all tokens} & \shortstack{Decoder input\\all tokens} & \shortstack{Decoder output\\all tokens} & \shortstack{Service (s)\\primary/all} & \shortstack{p95 (ms)\\primary/all}\\
\midrule
Full (one generation) & 1,110/1,117 & 0 & 2,404,792 & 36,551 & 827.473/830.474 & 1889.20/1874.83\\
\Native{} & 1,110/1,117 & 0 & 2,404,792 & 8,936 & 210.711/212.129 & 217.72/217.70\\
\Prefill{} & 1,110/1,117 & 0 & 2,412,611 & 1,117 & 47.499/47.804 & 75.48/75.44\\
Encoder, DEV threshold & 0/0 & 2,391,954 & 0 & 0 & 23.141/23.299 & 43.17/43.11\\
Abs0 & 216/217 & 2,391,954 & 671,201 & 217 & 31.559/31.747 & 75.11/74.82\\
Centered10 & 102/103 & 2,391,954 & 341,690 & 103 & 26.437/26.631 & 53.80/54.01\\
Centered25 & 250/253 & 2,391,954 & 833,150 & 253 & 34.199/34.484 & 94.10/93.88\\
Centered50 & 413/417 & 2,391,954 & 1,286,568 & 417 & 40.395/40.709 & 105.89/105.30\\
Hash10 & 102/103 & 2,391,954 & 301,249 & 103 & 26.277/26.471 & 53.15/53.62\\
Hash25 & 250/253 & 2,391,954 & 769,050 & 253 & 32.983/33.268 & 90.36/89.66\\
Hash50 & 413/417 & 2,391,954 & 1,232,347 & 417 & 40.186/40.529 & 104.25/104.06\\
\bottomrule
\end{tabular}
\caption{Complete fresh measured costs. Tokens cover the all 1,117 measured requests of that policy; primary/all entries are overlapping views of the same run. Raw/official full and native/calibrated encoder views share their respective row. Probe, warmup and cold/diagnostic costs appear separately in Table~\ref{tab:modern-phases-vnine}. Descriptive p95 is not a production-tail estimate.}
\label{tab:modern-cost-vnine}
\end{table*}

\Prefill{} improves its primary median/p95 service to 36.853/75.482\,ms
versus native-eight and full, but the routed policy does not dominate it
on tails: Centered25 p95 is 94.102\,ms. Centered50's p95 is 105.891\,ms.
The complete table retains these adverse values rather than inferring tails
from average call reduction.

Primary native-eight and prefill each consume exactly 2,391,172 logical
decoder input-plus-output tokens: prefill transfers 7,770 tokens from output
to input. Across all responses the equal total is 2,413,728, with 7,819
transferred tokens. In contrast, primary Centered25 sums to 3,190,274 logical
positions across both models, 33.4188\% above always-prefill's decoder
positions despite its lower measured service. Models have different sizes,
so this unweighted sum is not a FLOP comparison; neither call nor token
count establishes allocated-GPU, API-price, or energy savings.

\begin{table*}[t]
\centering\small
\begin{tabular}{lrrrr}
\toprule
Primary contrast & $\Delta$ F1 (pp) & Component 95\% interval & Service ratio & Component 95\% interval\\
\midrule
\Prefill{} $-$ full official & 0.0204 & $[-0.3364,0.4008]$ & 0.057403 & $[0.053997,0.061365]$\\
\Prefill{} $-$ full raw & -0.0770 & $[-0.3627,0.2120]$ & 0.057403 & $[0.053997,0.061365]$\\
\Prefill{} $-$ \Native{} & -0.0770 & $[-0.3627,0.2120]$ & 0.225423 & $[0.218730,0.234101]$\\
Centered10 $-$ Hash10 & 2.2926 & $[1.0647,3.5963]$ & 1.006095 & $[0.957551,1.058292]$\\
Centered25 $-$ Hash25 & 3.6034 & $[2.1127,5.2207]$ & 1.036865 & $[0.972890,1.110121]$\\
Centered50 $-$ Hash50 & 4.9559 & $[3.3058,6.5479]$ & 1.005203 & $[0.959123,1.053410]$\\
Centered25 $-$ Abs0 & -0.6473 & $[-1.9582,0.6210]$ & 1.083651 & $[1.028054,1.148355]$\\
Centered25 $-$ \Prefill{} & -4.0956 & $[-5.6205,-2.6624]$ & 0.719991 & $[0.677906,0.769094]$\\
Centered25 $-$ encoder & 5.3627 & $[3.8076,6.9942]$ & 1.477875 & $[1.394933,1.570663]$\\
\bottomrule
\end{tabular}
\caption{All prespecified primary contrasts. Ratios divide the first policy by the second. Intervals use 2,000 shared whole-component resamples, seed 20261025, conditional on fixed methods, actual outputs/routes, and one host/pass. They do not refit DEV calibration or estimate repeated scheduling variation.}
\label{tab:modern-contrasts-vnine}
\end{table*}

\paragraph{Source heterogeneity, with every source retained.}
The pooled primary contains 687 README responses. Tables~\ref{tab:modern-code-source-vnine}--\ref{tab:modern-readme-source-vnine}
therefore show all source-specific precision/recall/F1 points, not just the
pooled score. Code and README have no exposure-linked exclusions, so their
primary and complete source evaluations are identical. The tool table gives
both the primary 233 and complete 240 responses. Table~\ref{tab:modern-source-cost-vnine}
retains source-specific calls and complete warm service.

The centered primary policy is not uniformly better than Abs0: code F1 is
80.8081 versus 82.1622, README 94.9153 versus 96.0699, and tool 84.5815
versus 82.2967. Centered25 uses more calls than Abs0 in each source. Its
code service (11.0650\,s) is already close to always-prefill (11.1453\,s).
Thus a pooled gain over hash does not establish a universal routing upgrade.
The predeclared Centered50 point has higher observed tool F1 than
always-prefill, while its pooled F1 remains lower; that source-specific
observation is reported without a post-hoc policy or source choice.

\begin{table}[t]
\centering\small
\begin{tabular}{lrrr}
\toprule
Readout/policy & Precision & Recall & F1\\
\midrule
Full, official & 89.5349 & 81.0526 & 85.0829\\
Full, raw & 89.6552 & 82.1053 & 85.7143\\
\Native{} & 89.6552 & 82.1053 & 85.7143\\
\Prefill{} & 88.7640 & 83.1579 & 85.8696\\
Encoder, DEV threshold & 66.1290 & 86.3158 & 74.8858\\
Encoder, native zero & 75.0000 & 75.7895 & 75.3927\\
Abs0 & 84.4444 & 80.0000 & 82.1622\\
Centered10 & 71.3043 & 86.3158 & 78.0952\\
Centered25 & 77.6699 & 84.2105 & 80.8081\\
Centered50 & 84.3750 & 85.2632 & 84.8168\\
Hash10 & 66.3934 & 85.2632 & 74.6544\\
Hash25 & 70.7965 & 84.2105 & 76.9231\\
Hash50 & 76.7677 & 80.0000 & 78.3505\\
\bottomrule
\end{tabular}
\caption{Code-agent source, all 190 responses/176 components, including 95 released positives. Primary and complete populations are identical; all readouts and routed budgets remain. Values are percentages.}
\label{tab:modern-code-source-vnine}
\end{table}

\begin{table*}[t]
\centering\small
\begin{tabular}{lrrrrrr}
\toprule
Readout/policy & Primary $P$ & Primary $R$ & Primary F1 & All $P$ & All $R$ & All F1\\
\midrule
Full, official & 95.0000 & 81.8966 & 87.9630 & 95.0980 & 80.8333 & 87.3874\\
Full, raw & 95.0000 & 81.8966 & 87.9630 & 95.0980 & 80.8333 & 87.3874\\
\Native{} & 95.0000 & 81.8966 & 87.9630 & 95.0980 & 80.8333 & 87.3874\\
\Prefill{} & 95.0000 & 81.8966 & 87.9630 & 95.0980 & 80.8333 & 87.3874\\
Encoder, DEV threshold & 78.1513 & 80.1724 & 79.1489 & 78.6885 & 80.0000 & 79.3388\\
Encoder, native zero & 81.1881 & 70.6897 & 75.5760 & 81.7308 & 70.8333 & 75.8929\\
Abs0 & 92.4731 & 74.1379 & 82.2967 & 92.6316 & 73.3333 & 81.8605\\
Centered10 & 83.1858 & 81.0345 & 82.0961 & 83.6207 & 80.8333 & 82.2034\\
Centered25 & 86.4865 & 82.7586 & 84.5815 & 86.8421 & 82.5000 & 84.6154\\
Centered50 & 93.3962 & 85.3448 & 89.1892 & 93.5185 & 84.1667 & 88.5965\\
Hash10 & 80.7018 & 79.3103 & 80.0000 & 81.1966 & 79.1667 & 80.1688\\
Hash25 & 85.0467 & 78.4483 & 81.6143 & 85.4545 & 78.3333 & 81.7391\\
Hash50 & 89.4231 & 80.1724 & 84.5455 & 89.7196 & 80.0000 & 84.5815\\
\bottomrule
\end{tabular}
\caption{Tool source: fixed primary 233/217 components and complete 240/222. All sources and all seven linked tool responses are preserved; these populations overlap. Values are percentages.}
\label{tab:modern-tool-source-vnine}
\end{table*}

\begin{table}[t]
\centering\small
\begin{tabular}{lrrr}
\toprule
Readout/policy & Precision & Recall & F1\\
\midrule
Full, official & 99.1228 & 98.8338 & 98.9781\\
Full, raw & 99.1228 & 98.8338 & 98.9781\\
\Native{} & 99.1228 & 98.8338 & 98.9781\\
\Prefill{} & 98.8338 & 98.8338 & 98.8338\\
Encoder, DEV threshold & 84.3990 & 96.2099 & 89.9183\\
Encoder, native zero & 90.0850 & 92.7114 & 91.3793\\
Abs0 & 95.9302 & 96.2099 & 96.0699\\
Centered10 & 88.5942 & 97.3761 & 92.7778\\
Centered25 & 92.0548 & 97.9592 & 94.9153\\
Centered50 & 96.2751 & 97.9592 & 97.1098\\
Hash10 & 85.7513 & 96.5015 & 90.8093\\
Hash25 & 86.4230 & 96.5015 & 91.1846\\
Hash50 & 88.2979 & 96.7930 & 92.3505\\
\bottomrule
\end{tabular}
\caption{README source, all 687 responses/246 components, including 343 released positives. Primary and complete populations are identical; all readouts and routed budgets remain. Values are percentages.}
\label{tab:modern-readme-source-vnine}
\end{table}

\begin{table*}[t]
\centering\small
\begin{tabular}{lrrrrrr}
\toprule
Physical policy & Code $K$ & Code (s) & \shortstack{Tool $K$\\primary/all} & \shortstack{Tool service (s)\\primary/all} & README $K$ & README (s)\\
\midrule
Full (one generation) & 190 & 223.420 & 233/240 & 138.642/141.643 & 687 & 465.410\\
\Native{} & 190 & 38.538 & 233/240 & 46.150/47.568 & 687 & 126.023\\
\Prefill{} & 190 & 11.145 & 233/240 & 12.297/12.602 & 687 & 24.057\\
Encoder, DEV threshold & 0 & 6.504 & 0/0 & 5.729/5.887 & 0 & 10.908\\
Abs0 & 64 & 9.849 & 62/63 & 8.523/8.712 & 90 & 13.187\\
Centered10 & 30 & 8.012 & 29/30 & 6.966/7.159 & 43 & 11.459\\
Centered25 & 71 & 11.065 & 78/81 & 9.648/9.933 & 101 & 13.486\\
Centered50 & 116 & 12.893 & 124/128 & 11.750/12.063 & 173 & 15.753\\
Hash10 & 30 & 7.922 & 29/30 & 6.804/6.998 & 43 & 11.551\\
Hash25 & 71 & 10.404 & 78/81 & 9.232/9.518 & 101 & 13.347\\
Hash50 & 116 & 13.035 & 124/128 & 11.389/11.731 & 173 & 15.763\\
\bottomrule
\end{tabular}
\caption{Source-specific actual decoder calls and complete warm services. Code/README primary and complete source populations coincide. Tokens and all stage clocks remain in the aggregate artifacts; the pooled table does not conceal source-dependent costs.}
\label{tab:modern-source-cost-vnine}
\end{table*}

\paragraph{Prospectively defined no-truncation secondary.}
The encoder truncates 131 contexts in all 1,117 responses and truncates no
answer. The fixed primary no-encoder-truncation subset contains 979
responses/514 components and 485 released positives. Table~\ref{tab:modern-no-trunc-vnine}
reports every readout and budget there. Prefill F1 is 95.4879 versus the
calibrated encoder's 86.3724 and native encoder's 87.2951, so the inherited
capability gap is not explained only by the encoder context limit.
Centered25 remains below Abs0 (91.6583 versus 92.3400), with more calls and
service. Hash budgets were matched for source/fold/exposure strata on the
predeclared populations, not subsequently for this length subset; e.g.,
Centered25 has 191 versus Hash25's198 calls here. These subset contrasts
therefore carry their actual budgets and are not described as exactly
matched-call comparisons. The subset never replaces the full primary.

\begin{table*}[t]
\centering\small
\begin{tabular}{lrrrrrr}
\toprule
Readout/policy & Precision & Recall & F1 & 2B calls & Service (s) & p95 (ms)\\
\midrule
Full, official & 97.4304 & 93.8144 & 95.5882 & 979 & 684.6159 & 1742.99\\
Full, raw & 97.4304 & 93.8144 & 95.5882 & 979 & 684.6159 & 1742.99\\
\Native{} & 97.4304 & 93.8144 & 95.5882 & 979 & 182.6017 & 211.01\\
\Prefill{} & 97.2222 & 93.8144 & 95.4879 & 979 & 37.1757 & 48.22\\
Encoder, DEV threshold & 80.7899 & 92.7835 & 86.3724 & 0 & 17.5275 & 28.41\\
Encoder, native zero & 86.7617 & 87.8351 & 87.2951 & 0 & 17.5275 & 28.41\\
Abs0 & 94.0171 & 90.7216 & 92.3400 & 174 & 22.9145 & 59.96\\
Centered10 & 85.3659 & 93.8144 & 89.3910 & 77 & 19.1291 & 48.14\\
Centered25 & 89.4118 & 94.0206 & 91.6583 & 191 & 24.1784 & 62.70\\
Centered50 & 94.3983 & 93.8144 & 94.1055 & 329 & 28.4857 & 70.21\\
Hash10 & 82.4176 & 92.7835 & 87.2939 & 83 & 19.5099 & 50.74\\
Hash25 & 84.1808 & 92.1649 & 87.9921 & 198 & 23.6638 & 62.20\\
Hash50 & 87.1595 & 92.3711 & 89.6897 & 334 & 28.6980 & 70.05\\
\bottomrule
\end{tabular}
\caption{All prespecified readouts and budgets on the source-only no-encoder-truncation primary subset (979/514). Full aliases and encoder threshold views reuse identical physical costs. The subset is not a separate inference experiment.}
\label{tab:modern-no-trunc-vnine}
\end{table*}

\paragraph{Complete execution, costs, and validator repair.}
All 13,408 planned request records complete: 12,287 measured requests,
1,117 encoder-probe requests, and four warmups. Actual calls total 10,054
encoder forwards and 5,117 decoder generations, including warmup. No route
drift breaks exact hash matching on the primary or complete populations.
The actual-input/output and sampler audits retain every submitted call;
all 5,117 decoder calls pass the independent CPU ID audit. The auditor reports
78.995\,s internally; the collector measures 80.645\,s for its complete audit
process. These nested clocks are not added. The final local scorer takes
32.004 CPU seconds and adds no inference. Separate pre-submission CPU work
costs 34.894\,s for encoder-only qualification and 118.961\,s for the native
preflight worker, without model weights, GPU allocation, or labels. The
latter includes its own nested 100.256\,s token/ID preflight. These external
CPU clocks are distinct from the actual GPU worker's 103.922\,s token/ID
qualification in Table~\ref{tab:modern-phases-vnine}, which remains inside
the complete allocation. The table retains that allocation's other stage
boundaries.

The first frozen local validator refused before reference access because it
looked up the worker's full-repeat guard under a mismatched configuration
key. The original refusal and code remain. A separately recorded v2
validator changes only that key lookup to the worker's recorded field;
it retains the original cohort, model outputs, routes, metrics, bootstrap,
and costs. There is no new inference, regenerated arm, or method repair.
It passes the same independent pre-reference audit before scoring.

\begin{table}[t]
\centering\small
\begin{tabular}{lr}
\toprule
Stage & Seconds\\
\midrule
Decoder initialization & 40.204\\
Encoder initialization & 1.711\\
Within-worker token/ID qualification & 103.922\\
Asset verification & 6.017\\
Probe service & 24.056\\
Probe diagnostics & 5.225\\
Four warmups, service & 3.972\\
Four warmups, diagnostics & 0.020\\
Measured-arm diagnostics & 93.166\\
Route planning & 0.173\\
Window primitives & 0.221\\
Own-engine shutdown & 3.808\\
Complete worker wall & 1903.174\\
Actual main allocation (GPU-s) & 1,907\\
\bottomrule
\end{tabular}
\caption{Paid modern confirmation stages. Policy warm clocks are separately reported in Table~\ref{tab:modern-cost-vnine}; the whole worker/main allocation includes all phases, remaining preparation/final-flush/cleanup work and idle time. Nested clocks are not added again, and common costs are not attributed selectively to a policy. Post-release CPU audit/scoring are outside the GPU allocation.}
\label{tab:modern-phases-vnine}
\end{table}

The complete worker lasts 1,903.174\,s; terminal main-job accounting is
one allocated GPU for 1,907 actual seconds, counted once. Across all recorded
phases the audit retains 21,529,080 encoder input positions,12,662,380
decoder input positions, and 48,386 generated tokens. These include paid
probe and warmup; the per-policy token table covers measured requests only.
The complete project at this confirmation's closure contains 115 terminal
allocations and 86,597 actual GPU-seconds (24.0547 GPU-hours), including
abandoned directions, startup, failures and idle work. API expenditure is
USD~0. Warm reductions are not multiplied into this whole-job or project
cost as a billing claim.

The original native-readout contribution is independently supported here
under common fresh preparation. The centered allocation remains a useful
point against equal-call hash but fails its stronger Abs0 transfer contrast;
all budgets, source differences, tail costs, and this negative result are
retained. No new threshold/prompt/seed/source rescue follows this opened
confirmation. The signals remain released-reference response alarms and do
not certify factual truth, complete localization, or successful execution.

\section{Conditional prefix stability and finite-sample accounting}
\label{app:native_branch_theory}

\paragraph{Conditional causal-prefix lemma.}
Let $g(p)$ be a deterministic greedy next-token rule for complete token prefix
$p$, including a fixed rule for breaking ties. Suppose an ordinary native run
from $p$ generates syntax prefix $s$. If a prefilled run starts from the identical
complete prefix $p\mathbin{\|}s$ and evaluates the same next-token function,
with identical inference arithmetic and relevant state, its continuation equals
the ordinary run's continuation. The first next-token argument is identical;
induction applies after each common next token. Therefore a parser determined
by this prefix and next token returns the same branch under these premises.
This is a conditional identity, not a guarantee that two GPU execution paths
compute identical logits or that a generated output is factually correct.

\paragraph{A sufficient numerical condition.}
At an identical complete prefix, let $\ell$ and $\widetilde\ell$ be two finite
all-vocabulary logit vectors with
$\lVert\ell-\widetilde\ell\rVert_\infty\leq\epsilon$.
If the original top-one versus runner-up gap exceeds $2\epsilon$, the top token
cannot change: the top logit can fall by at most $\epsilon$, and every competing
logit can rise by at most $\epsilon$. This condition must hold at each relevant
continuation prefix for a longer common trajectory. Our closed traces do not
contain all-vocabulary logits and provide neither an empirical $\epsilon$ nor
a numerical certificate from this condition. Token agreement alone does not
estimate either quantity.

\paragraph{Exact F1-change bounds without individual labels.}
Consider $n$ responses with $P$ positive reference labels and Boolean positive
prediction sets $A$ and $B$. Partition the responses into $C=A\cap B$,
$E_A=A\setminus B$, $E_B=B\setminus A$, and $O=(A\cup B)^c$, with respective
sizes $m_C,m_A,m_B,m_O$. Let $x_j$ be the number of positive labels in group $j$.
These integers satisfy $0\leq x_j\leq m_j$ and $\sum_jx_j=P$.
Define $\alpha=2/(P+|A|)$ and $\beta=2/(P+|B|)$, assigning coefficient zero
when its denominator is zero, consistently with zero F1 for that degenerate
case. Then
\begin{equation}
 F_1(B)-F_1(A)
 = (\beta-\alpha)x_C-\alpha x_A+\beta x_B.
\end{equation}
Thus the coefficients for $(C,E_A,E_B,O)$ are
$(\beta-\alpha,-\alpha,\beta,0)$. The exact maximum follows by filling $P$
integer label units into groups in descending coefficient order, up to each
capacity; ascending order gives the exact minimum. If a lower-coefficient group
contains a unit while a higher-coefficient group has capacity, transferring that
unit cannot decrease the objective, proving the extremal construction. Every
feasible integer group allocation is realized by some assignment of labels to
the responses, so both endpoints are attainable. The procedure uses only
$n,P,|A|,|B|$, and $|A\cap B|$, without individual reference labels.

This bound is finite-sample accounting rather than a confidence interval or a
population quality-equivalence test. Individual released labels are used
separately to attribute observed true-positive and false-positive changes and
to locate the observed difference inside the bound. The analysis retains every
previously fixed population. Unknown outputs are reported and do not trigger
case deletion to obtain a Boolean-bound result. Few changing predictions can
mechanically limit F1 changes; this does not establish evidence dependence,
entailment, real-error accuracy, or newly acquired model ability.

\paragraph{Observed branch and mapping attribution.}
This additional analysis reuses the closed modern confirmation outputs;
it adds no model calls. All nine previously fixed populations remain.
In the primary 1,110 responses, the full generation's first-eight-token
branch and its final raw-list alarm agree on every response.
The official offset mapper separately removes one true positive, changing
F1 by $-0.097456$ points. Native-eight and syntax-prefill differ on only
three responses, all negative-to-positive: one additional true positive
and two additional false positives. Their observed F1 difference is
$-0.077032$ points. Given only the population positive count and these
prediction-set sizes/overlap, the exact attainable range is
$[-0.269866,+0.552486]$ points. This explains why few altered alarms can
bound a small aggregate quality change. Evidence conditioning is examined
separately in the fixed intervention assay; the paired intervals describe
this finite prediction comparison. The original code-agent
trajectory differences and every adverse transfer result are unchanged.

\section{Evidence interventions and balanced runtime controls}
\label{app:native_validity}

\paragraph{Upstream development exposure and reference provenance.}
The additional modern cohort is project-unopened validation, not a clean
external test. In Section~6.1, \citet{grounding} explicitly select the
mmBERT checkpoint by development hallucinated-token F1 on the upstream
unified split. The encoder is therefore upstream development-exposed on
this partition, even though our cohort selection, source closure, and
comparison were fixed before our project inspected its responses or
labels. The cited paper does not specify an analogous checkpoint-selection
rule for Qwen3.5-2B, and we do not extend the encoder attribution to that
generator. Section~8 also distinguishes the model-assisted author review
of code-agent TEST from automatically checked injection labels in
development and the non-code test sources. Released span-list presence
therefore defines a common, reproducible quality target for every readout
under this annotation provenance.

\paragraph{Scope of the controlled evidence-balance assay.}
Our separate construction crosses two recorded target-field values with
two fixed literal-value responses in each of 24 tool-return families,
retaining all 96 combinations. Each exact response and each exact context
occurs once with each label within its family. Excluding opaque identifiers,
a classifier that sees only the response or only the context has at most
50\% empirical accuracy on this complete balanced pool. Labels are checked
by parsing the rendered JSON target field and the literal in the rendered
response, rather than by assuming a real tool was executed. This removes
one-sided label imbalance and permits a bounded test of whether a frozen
readout changes correctly with the supplied fact. The measured readout decisions are evaluated against this literal-consistency
target. The complete crossed pool identifies whether decisions respond to
the supplied value; its unit of construction is the tool-return family,
with four dependent inputs per family.

\paragraph{Prospective evidence-intervention protocol.}
Four fixed fixtures in each of six developer-tool domains supply 24
families: failed-test count, command exit code, file count, configuration
scalar, artifact path, and a Git field. A family has two JSON target-field
values and two responses asserting the corresponding literal values.
Crossing them yields two supported and two unsupported responses. The
request, distractor fields, and response wording remain fixed when the
target value is changed; complete answers remain fixed within the
evidence-flip pairs. All 24 quartets, 48 fixed-answer pairs, and 96 inputs
remain, including any failures. Opaque identifiers are not supplied to
the detector. The unchanged native wrapper receives the complete supplied
record and answer, without the local evaluator's labels or latent states.
No model output was inspected to select a fixture, prefix, prompt,
threshold, or additional sample.

The frozen full, native-eight, and syntax-prefill readouts each run once
on every input, with arm order rotated by input index, under the
single-sequence engine. Full raw-list and official-offset alarms reuse
the same full generation. Primary summaries are the number of completely
correct quartets and the number of fixed-answer pairs for which both
evidence states receive the correct opposite judgments. Accuracy, F1,
confusion, every domain, invalid-output sensitivities, and readout
disagreements accompany those family/pair summaries. These deliberately
constructed families are the measurement units; they are not 96 randomly
sampled independent natural reports.

\paragraph{Fixed source-only runtime selection.}
The separate timing pool reuses 24 previously closed natural responses
without reading their quality labels, predictions, or measured times.
Within each of the three modern source types, eligible project-unlinked
rows are ranked by the untruncated encoder-pair token count and identifier.
Each length quartile contributes two fixed-salt hash-selected rows in
distinct source components. All 24 are retained and interleaved into three
fixed batches of eight. The selection length is a source-only stratifier,
not the decoder's token count. Their prior use precludes a new natural
quality-confirmation claim.

For each batch, all six permutations of full/native-eight/prefill execute
under each of two sequential engines with maximum active sequences one
and eight. A separately paid prefill call on the batch's first row
immediately precedes every permutation; this fixes the first arm's
predecessor while all permutations balance positions and within-order
predecessors. Each engine receives its own three first-row warmups.
Both explicitly use 32,768 batched tokens and chunked prefilling, BF16,
eager execution, seed zero, and no prefix cache. This new robustness
setting does not reproduce the earlier default scheduler or certify its
exact speed percentage. Engine one precedes engine eight in one job,
so cross-engine comparisons retain time/order confounding.

\paragraph{Complete clocks and accounting.}
The complete plan contains 438 generate calls and 1,194 submitted
sequences: 288 evidence-intervention sequences, 864 natural timing
sequences, 36 conditioners, and six warmups. Each of the 108 actual
eight-response timing calls has one whole-batch service clock beginning
before fresh required input preparation and ending after generation,
parsing, and the eight minimal alarm writes with flush/fsync calls.
Dividing this clock by eight is amortized service, not individual latency.
We report every engine/order/batch and descriptive ratio ranges. The
three source batches with six repeated orders are not 108 independent
natural samples or a production-tail estimate. Source qualification,
both initializations, complete actual-ID diagnostics, conditioners,
warmups, shutdown, failures, and allocation idle time remain paid;
post-release CPU audit and scoring are separate. No warm-service ratio
is assigned as an allocated-GPU, monetary, or energy saving.

\paragraph{Evidence-conditioned results.}
All four readouts---full official-offset presence, full raw-list presence,
\Native{}, and \Prefill{}---classify all 96 constructed inputs correctly.
For each readout, all 24 quartets are completely correct and all 48
fixed-answer pairs receive the correct opposite decisions after the
evidence change. Table~\ref{tab:evidence_domains_vten} reports every
prespecified domain. Each readout has 48 true positives, 48 true negatives,
zero false positives, and zero false negatives; accuracy, precision,
recall, and F1 are all 100\%. Invalid-as-negative, invalid-as-positive,
and strict-invalid-always-wrong summaries coincide because there are no
invalid outputs. Full official and raw alarms reuse one generation, and
all six pairwise readout comparisons agree on all 96 inputs.

This complete finite-pool result rules out a deterministic answer-only or
context-only decision rule for these constructed records: without opaque
identifiers, either feature alone permits at most 50\% accuracy and no
fully correct quartet. It supplies evidence that the shortened readouts
retain the frozen detector's literal evidence-conditioned behavior on
this assay. The 96 crossed inputs form 24 deliberately constructed
families, so the result is a complete-pool measurement of literal
conditionality with family dependence retained in its interpretation.

\begin{table}[t]
\centering\small
\begin{tabular}{lrrr}
\toprule
Domain & Correct & Quartets & Pairs \\
\midrule
Failed-test count & 16/16 & 4/4 & 8/8 \\
Command exit code & 16/16 & 4/4 & 8/8 \\
File count & 16/16 & 4/4 & 8/8 \\
Configuration scalar & 16/16 & 4/4 & 8/8 \\
Artifact path & 16/16 & 4/4 & 8/8 \\
Git field & 16/16 & 4/4 & 8/8 \\
\midrule
All & 96/96 & 24/24 & 48/48 \\
\bottomrule
\end{tabular}
\caption{Complete evidence-intervention results. Every entry applies
separately to full official, full raw, native-eight, and prefill-one.
``Quartets'' requires all four decisions correct; ``Pairs'' requires
both fixed-answer evidence states correct, with opposite decisions.
All inputs and all domains are retained.}
\label{tab:evidence_domains_vten}
\end{table}

\paragraph{Balanced-order runtime results.}
Table~\ref{tab:runtime_totals_vten} gives all three arms under each engine.
Every arm has 18 actual eight-response timing calls per engine: three
fixed source batches repeated under all six orders, totaling 144 submitted
sequences. Whole-service totals include fresh preparation, generation,
parsing, and minimal alarm writes. Prefill/full service ratios are
0.047143 and 0.074726 under maximum active sequences one and eight,
respectively; prefill/native-eight ratios are 0.230934 and 0.618659.
Thus the measured prefill reduction against full is 95.29\% and 92.53\%,
while the reduction against native-eight is 76.91\% and 38.13\%.
The latter benefit is smaller in the eight-active-sequence setting.
These within-engine contrasts retain all orders rather than selecting
the fastest ordering.

\begin{table*}[t]
\centering\small
\begin{tabular}{rlrrrrr}
\toprule
Active seq. & Arm & Service (s) & Generate (s) & Prepare (s) & Input tokens & Output tokens \\
\midrule
1 & Full & 144.0244 & 142.7053 & 1.0859 & 471,930 & 6,024 \\
1 & Native-eight & 29.4013 & 28.1241 & 1.0847 & 471,930 & 1,152 \\
1 & Prefill-one & 6.7898 & 5.5692 & 1.0492 & 472,938 & 144 \\
\midrule
8 & Full & 58.8752 & 57.7138 & 0.9865 & 471,930 & 6,024 \\
8 & Native-eight & 7.1114 & 5.9506 & 0.9983 & 471,930 & 1,152 \\
8 & Prefill-one & 4.3995 & 3.2270 & 1.0063 & 472,938 & 144 \\
\bottomrule
\end{tabular}
\caption{Complete natural-input runtime totals for the new robustness
setting. Each row covers the same 24 fixed responses under six balanced
orders. Prefill input includes its seven syntax tokens on every submitted
sequence. Preparation and generation are components of service; parse and
minimal-write clocks are also included. These are whole-batch clocks,
not individual-response latency. Natural quality labels are not read.}
\label{tab:runtime_totals_vten}
\end{table*}

Table~\ref{tab:runtime_orders_vten} reports all six orders for both
engines. Each ratio there divides the summed services of the same three
source batches. Separately, the ranges of the 18 paired batch ratios
(three batches by six repeated orders) are as follows. With one active
sequence, prefill/full ranges from 0.034517 to 0.055414,
prefill/native-eight from 0.180695 to 0.312688, and native-eight/full
from 0.177218 to 0.251142. With eight active sequences, the corresponding
ranges are 0.030627--0.121499, 0.445771--0.788292, and
0.068707--0.164406. They describe these fixed repetitions; neither the
ranges nor the 108 timed calls are independent-sample uncertainty
intervals or production-tail estimates.

\begin{table}[t]
\centering\small
\begin{tabular}{rcrrr}
\toprule
Active & Order & P/F & P/N & N/F \\
\midrule
1 & FNP & .049383 & .248561 & .198676 \\
1 & FPN & .047463 & .227374 & .208744 \\
1 & NFP & .047096 & .228481 & .206128 \\
1 & NPF & .046446 & .229299 & .202556 \\
1 & PFN & .046860 & .228397 & .205169 \\
1 & PNF & .045666 & .224303 & .203591 \\
\midrule
8 & FNP & .070717 & .607288 & .116448 \\
8 & FPN & .078201 & .647498 & .120774 \\
8 & NFP & .073836 & .596061 & .123874 \\
8 & NPF & .075123 & .611470 & .122856 \\
8 & PFN & .075919 & .632036 & .120118 \\
8 & PNF & .074802 & .618622 & .120917 \\
\bottomrule
\end{tabular}
\caption{All prespecified order contrasts. F, N, and P denote full,
native-eight, and prefill-one. Ratios use complete service summed over
the three fixed batches for that engine/order, rather than the mean of
individual batch ratios. ``Active'' is the engine's maximum active
sequence count.}
\label{tab:runtime_orders_vten}
\end{table}

For every one of the 24 natural rows and each arm, the 12 observations
across both engines and all orders have one unique generated token-ID
sequence and one unique decision. This is an actual-output stability
check, not a new natural quality measurement. All submitted outputs are
valid. No full generation reaches its 1,024-token cap. Native-eight
reaches its intentional eight-token limit on all 386 submitted sequences,
and prefill-one reaches its intentional one-token limit on all 422,
including intervention, timing, warmup, and conditioner phases as
applicable. These planned short-output cap hits are retained rather
than interpreted as full-output failures.

The engines run sequentially on one host in one allocation, with the
one-active-sequence engine first. Consequently, a cross-engine change
cannot be attributed solely to concurrency: elapsed time, cache state,
and engine order remain confounded. Both engines explicitly use chunked
prefilling and a 32,768 batched-token budget, which differs from the
earlier scheduler defaults. The assay supports robustness of a useful
within-generator operating point under this additional setting, not
exact replication of the earlier percentage, production latency,
quality equality, or fieldwide SOTA.

\paragraph{Complete execution, validation, and cost.}
All 438 generate calls complete, covering 1,194 submitted sequences:
288 intervention, 864 natural timing, 36 conditioner, and six warmup
sequences. Their respective complete service totals are 70.483,
250.602, 1.448, and 7.264\,s. Total input/output tokens over all phases
are 3,059,522/17,633. Engine initialization takes 150.423/39.677\,s
and shutdown 1.944/3.236\,s for active-sequence limits one/eight;
actual-ID diagnostics total 10.822\,s. Asset verification and source
tokenization preflight take 5.776 and 5.381\,s. These component clocks
are included in the 590.746\,s worker wall time, not added to it as
separate allocations. The main allocation is one GPU for 593 elapsed
seconds, counted once, including setup, failures if any, and idle time.
The independent post-job CPU ID auditor verifies all 1,194 sequences
in 23.283\,s internally; its collector process takes 24.897\,s.
A separate CPU-only pre-submission qualification takes 36.519\,s
without loading weights, constructing an engine, or generating outputs.
Local scoring takes 0.333 CPU seconds, with no new model calls or GPU
use. All audit checks pass and scoring reports zero issues. Warm-service
ratios are not allocation, billing, monetary, or energy-saving estimates.

The initial local scorer refuses before reading references because its
transport validator resolves fixed remote tokenizer/config paths to
nonexistent Mac assets. A pinned validation-only revision instead uses
the already collected independent remote CPU hash attestation for those
exact assets; all other pins still require local bytes. The refusal and
original scorer remain preserved. This changes no inference, method,
source, route, metric, or output, and copies no model/tokenizer assets
to the Mac. The completed score and aggregate tables refer to that
strictly bounded transport repair.

\section{Matched-contract and optimized-execution controls}
\label{app:increment_controls_vfifteen}

Two additional controls distinguish the usefulness of the native readout from
that of its established inference primitives. The first preserves the verifier's
semantic instructions while replacing its output contract with a direct Boolean
answer. The second tests whether moving syntax into prefill remains useful when
the decoder uses its default compiled execution path. These controls reuse
exposed development or label-blind timing inputs; they do not add an independent
modern test cohort. All comparisons use the same frozen checkpoint and tokenizer
revision as the focal readout.

\subsection{A semantically matched binary contract}
\label{app:binary_contract_vfifteen}

The complete Data2txt development source contains 1,026 responses in 171
six-response source components, with 715 positive and 311 negative references.
The source was already exposed during development and originates from official
TRAIN, so this comparison diagnoses a fixed interface choice rather than
providing fresh confirmation. Every selected response is retained. References
are evaluator-only and are not passed to either prompt construction or decoding.

The original opening definition of a hallucinated span, every category and
subcategory definition, the evidence, answer, and user wrapper are byte-identical
across arms. We replace the native action to quote and classify spans with the
fixed instruction to determine whether any such span exists. The binary output
contract requests a JSON object with \code{has\_annotated\_error} set to
\code{true} or \code{false}. Its supplied prefix ends immediately after that
key's colon. We compare native \Prefill{}, unrestricted binary continuation,
and masked binary continuation. Each generates one greedy token, with no
prompt, threshold, or seed search. The mask is a stronger format control than
unrestricted binary generation: it restricts the continuation to every legal
single-token Boolean spelling, without changing the verification definition.

The legal alphabet is determined by syntax, not prediction quality. We enumerate
all nonspecial vocabulary tokens, admit only ASCII JSON whitespace around
\code{true}/\code{false} with an optional object-closing brace, and verify
compositional decoding after the supplied prefix. The six accepted IDs are
804 (space--true), 867 (space--false), 1802 (true), 3721 (false),
34819 (tab--false), and 41393 (tab--true). No token in this vocabulary combines
an admitted Boolean with the closing brace. The binary prefix IDs are
$(4754,4497,62,3292,639,3960,763)$; the native seven IDs are unchanged.
Both binary parsing and masking use the complete six-ID alphabet.

\begin{table}[t]
\centering
\begin{tabular}{lrrrr}
\toprule
Readout & Prec. & Recall & F1 & Valid \\
\midrule
Native \Prefill{} & 94.03 & 94.69 & 94.36 & 100.00 \\
Binary, masked & 96.33 & 88.11 & 92.04 & 100.00 \\
Binary, free & 93.92 & 23.78 & 37.95 & 35.77 \\
\bottomrule
\end{tabular}
\caption{Matched-definition binary control on exposed DEV. All values are
percentages. Free binary has 659 invalid outputs: unknown positives count as
false negatives in F1, while unknown negatives remain explicit rather than
becoming true negatives. Its F1 is therefore accompanied by valid coverage;
the fully valid masked binary arm is the fixed-default contract control.}
\label{tab:binary_contract_vfifteen}
\end{table}

Table~\ref{tab:binary_contract_vfifteen} reports the corrected comparison.
Native \Prefill{} has 677 true positives, 43 false positives, 38 false negatives,
and 268 true negatives; masked binary has 630, 24, 85, and 287, respectively.
The fixed-default native F1 difference is 2.32 points, with a paired 95\% component-bootstrap
interval of $[1.10,3.59]$. We resample the 171 whole source components with
shared draws for both arms, using 2,000 replicates and seed 20261102. This
interval conditions on the exposed cohort and the fixed outputs. It is not an
independent-confirmation or equivalence interval. Among 70 disagreements,
native is correct and binary wrong in 49 cases; the reverse occurs in 21.
This default difference is a precision--recall tradeoff: masked binary has higher
precision, while native retains more positive references. It is not dominance
on every metric.

For unrestricted binary, 514 positive and 145 negative examples remain unknown.
The remaining counts are 170 true positives, 11 false positives, 31 valid false
negatives, and 155 true negatives. Strict all-item accuracy, counting every
unknown as wrong, is 31.68\%; the corresponding worst-case binary F1 bound is
32.66\%. We neither repair these outputs nor discard them. However, format
failures in this arm do not substitute for the fully valid masked comparison.

Each arm processes 33 batches of at most 32 responses, in original source
order, with the three arm orders rotated by batch index. One singleton warmup
per arm precedes the measurements. BF16 eager execution uses an active-sequence
cap of eight, maximum context 16,384, no prefix caching, and greedy seed zero.
There are 3,078 measured sequences plus three warmups. Fresh service includes
fresh tokenization, synchronized generation, parsing, and per-alarm minimal
synchronous writing. Native, masked binary, and free binary take 14.379,
14.661, and 14.481 seconds of measured service, respectively; their generation
times are 10.068, 10.090, and 10.018 seconds. Native input/output totals are
1,471,501/1,026; both binary arms use 1,462,267/1,026. All supplied syntax is
counted as input. These costs are comparable; the approximately 2\% service
difference is not the main efficiency result. Shared initialization takes
53.100 seconds, with qualification, warmup, and diagnostics separately retained.

This comparison describes one exposed-DEV operating point under a fixed,
semantics-matched contract. We do not infer intrinsic quality superiority over
calibrated binary readouts, or a causal need for span training.
The exact prompts, engine settings, token alphabet, confusion counts, uncertainty
specification, and phase costs accompany this appendix in the anonymous
aggregate artifact.

\subsection{Default compiled execution}
\label{app:optimized_runtime_vfifteen}

The runtime control reuses 24 previously qualified label-blind inputs in three
fixed batches of eight. Each batch is measured in both Native8--Prefill1 and
Prefill1--Native8 order. We run four separate processes, in the fixed order eager
with active-sequence cap one, optimized with cap one, optimized with cap eight,
and eager with cap eight. Each process has 96 measured sequences and two
singleton warmups, for 392 sequences in total. These are three fixed batches
with repeated orders, not six independent samples or a new quality population.

All processes share BF16 arithmetic, maximum context and batched-token budgets
of 32,768, chunked prefill enabled, prefix caching disabled, memory utilization
0.70, and seed zero. The requested intervention changes \code{enforce\_eager}
from true to false. The resulting default optimized execution uses the Inductor
backend, compilation mode 3, and full-and-piecewise CUDA Graphs. The captured
size sets are $\{1,2\}$ and $\{1,2,4,8,16\}$ for the two active-sequence caps.
Raw process logs, in addition to configuration snapshots, confirm compilation
and graph capture: the two optimized processes report 47.86 and 11.09 seconds
for compilation and one and two seconds for capture. This comparison measures
the default optimized path; it does not isolate the causal effect of CUDA Graphs
from the compiler, backend, or cache state.

\begin{table}[t]
\centering
\begin{tabular}{lrrrr}
\toprule
Execution & $S$ & Native8 & Prefill1 & Ratio \\
\midrule
Eager & 1 & 10.250 & 2.560 & .250 \\
Optimized & 1 & 3.152 & 2.535 & .804 \\
Eager & 8 & 2.820 & 1.554 & .551 \\
Optimized & 8 & 1.692 & 1.542 & .912 \\
\bottomrule
\end{tabular}
\caption{Fresh-service totals in seconds on the same timing source.
$S$ is the active-sequence cap, not the number of concurrent clients.
Ratio is the sum of Prefill1 service divided by the sum of Native8 service,
not the median of the six paired batch/order ratios. Initialization and warmup
are separate from these totals and included in complete allocation costs.}
\label{tab:optimized_runtime_vfifteen}
\end{table}

Table~\ref{tab:optimized_runtime_vfifteen} shows a smaller but positive aggregate
increment over native early stopping after optimization: Prefill1 reduces fresh
service by 19.57\% at cap one and 8.84\% at cap eight. The corresponding median
paired service ratios are .706 and .920; eager medians are .247 and .578.
These summaries answer different questions: the ratio of totals weights batches
by their Native8 duration, whereas the median gives each paired batch/order
ratio equal weight. We report both rather than substituting the more favorable
summary.

\begin{table}[t]
\centering
\begin{tabular}{lrr}
\toprule
Batch/order & $S=1$ & $S=8$ \\
\midrule
1 / Native first & 1.372 & .774 \\
1 / Prefill first & .628 & 1.053 \\
2 / Native first & .624 & 1.032 \\
2 / Prefill first & .691 & .937 \\
3 / Native first & .721 & .839 \\
3 / Prefill first & .785 & .903 \\
\bottomrule
\end{tabular}
\caption{Every optimized paired service ratio, Prefill1/Native8. Values above
one are retained: Prefill1 is slower in one of six batch/order measurements at
cap one and two of six at cap eight. Repeated orders are not independent
statistical samples. The aggregate artifact also retains all eager ratios.}
\label{tab:optimized_orders_vfifteen}
\end{table}

The benefit is not uniform across measurements
(Table~\ref{tab:optimized_orders_vfifteen}). Each arm has 48 measured sequences
per process. Native8 input/output totals are 157,310/384 and Prefill1 totals are
157,646/48, giving the same 157,694 logical tokens. There are no invalid outputs
or within-process branch disagreements. Across eager and optimized execution,
all 192 matched output-ID and decision comparisons agree. This is observed
agreement on the timing source, not a general finite-precision identity claim.

Initialization, including compilation or capture where applicable, takes
46.071, 88.540, 55.585, and 44.359 seconds in the stated process order. Warmups,
source qualification, diagnostics, and initialization are charged to the complete
allocation rather than selectively assigned to a favorable arm. The fresh-service
clock includes tokenization, synchronized generation, parsing, and minimal alarm
writing. It does not measure per-request response latency, concurrent-client
serving, a complete application workflow, or proportional GPU billing. The
smaller optimized increment also cautions against transferring an eager speedup
unchanged to another runtime configuration.

\subsection{Validity lineage and complete accounting}

Two implementation defects were identified and retained rather than converted
into apparent research gains. First, a CPU preflight assertion incorrectly
expected the native seven-ID prefix to include the list opener. The prefix
actually ends at the colon; the next generated token carries the list opener
and branch. Only this assertion was corrected before GPU inference, with no
change to the native prefix, prompt, or parser.

Second, the initial binary parser and mask admitted only bare true/false token
IDs. A syntax audit found 366 legal leading-space Boolean outputs that the
parser had rejected. That incomplete alphabet is not evidence of detection
failure. We retained its raw outputs and superseded its capability comparison
with the prospective, complete six-token grammar described above. The corrected
three-arm run uses the same semantic prompts, inputs, and native readout; no
label or performance ranking selects aliases. All corrected outputs, including
the remaining 659 unrestricted-format failures, are retained. The paper's
matched-contract comparison uses only the corrected result.

The first allocation, containing the superseded binary comparison and all four
runtime processes, consumes 577 GPU seconds; the corrected binary allocation
consumes 167. Both are terminal, with one allocated GPU each. Their combined
744 GPU seconds include startup, qualification, diagnostics, failed-control
work, and idle overhead. Complete project exploration therefore comprises
119 terminal allocations and 88,215 GPU seconds (24.5042 GPU hours), with zero
pending reservation and zero paid API expenditure. These allocation costs are
separate from service ratios and are not attributed to individual methods.

The separately maintained artifact archive contains \path{interface_controls.csv} for the
reported aggregates and
\path{interface_controls.json} for the exact
contracts, token spellings, schedules, metrics, all paired runtime ratios,
software versions, input/reference/code hashes, and complete cost scopes.
Neither artifact contains individual inputs, identifiers, reference labels, or
predictions. Raw instances and full execution evidence remain separately
preserved; aggregate reconstruction and source hashes do not replace them for
independent rerunning.

\section{Optimized execution on the complete workload}
\label{app:optimized_full_vseventeen}

The full-workload control asks whether the extra saving over native early
stopping survives both optimized execution and the complete set of 2,015
code-agent inputs. It extends the 24-input timing pilot; the checkpoint,
native instruction, seven-token prefix, and alarm parser remain fixed.
The primary comparison is the ratio of summed fresh-service times for
Prefill1 and Native8. No labels, output-based selection, prompt search,
or retries enter this timing measurement.

\subsection{Workload and execution}

The original input bytes are retained in their frozen order and partitioned
into 63 chunks: 62 of size 32 and one of size 31. Each chunk is processed in
both Native8--Prefill1 and Prefill1--Native8 order, with the initial order
alternating by chunk index. Thus each arm processes every input twice:
4,030 measured sequences per arm at each active-sequence cap. Each process
also has one singleton warmup per arm, giving 8,062 sequences including
warmups. The two processes run sequentially on the same H100 80GB allocation,
first with active-sequence cap $S=1$, then $S=8$. These are scheduler caps,
not numbers of independent clients. The two order repetitions reuse the same
inputs and are not independent quality samples.

Both processes use vLLM 0.17.1, BF16, greedy decoding with seed zero, context
and batched-token limits of 32,768, chunked prefill enabled, memory utilization
0.70, and prefix caching disabled. There are no masks or log-probability
requests in the timed calls. Default optimized execution uses Inductor
compilation and full-and-piecewise CUDA Graphs. Process logs confirm loading
compiled graphs and completing mixed prefill/decode and full-decode graph
capture; this conclusion does not rely on configuration flags alone.
Compilation caches are reused, so the experiment is not a cold-cache
compilation benchmark. The graph capture sets are $\{1,2\}$ for $S=1$ and
$\{1,2,4,8,16\}$ for $S=8$.

Every measured call freshly constructs and tokenizes its input, performs
synchronized generation, parses the alarm, and minimally writes it with
synchronous persistence. Their sum is the fresh-service clock. Qualification,
initialization, warmup, and the more extensive actual-token-ID diagnostic
serialization are timed separately. The first measured chunk remains in the
reported totals. Each arm has exactly 17,242,482 measured logical tokens at
each cap, across its two workload passes: Native8 uses 17,210,242 input and
32,240 output tokens; Prefill1 uses 17,238,452 input and 4,030 output tokens.
The supplied syntax is therefore charged as input, rather than counted as
free or deleted from the total.

\begin{table}[t]
\centering
\begin{tabular}{rrrrr}
\toprule
$S$ & Native8 (s) & Prefill1 (s) & Saving & Slower \\
\midrule
1 & 250.121 & 183.263 & 26.73\% & 0/126 \\
8 & 166.308 & 155.106 & 6.74\% & 3/126 \\
\bottomrule
\end{tabular}
\caption{Complete-workload optimized fresh service. Times sum \emph{two}
passes of all 2,015 inputs per arm; they are not single-pass times. Saving is
one minus the ratio of summed Prefill1 to Native8 service. Slower counts
paired chunk/order observations with a ratio above one.}
\label{tab:optimized_full_runtime_vseventeen}
\end{table}

\subsection{Service savings and complete costs}

Table~\ref{tab:optimized_full_runtime_vseventeen} shows a positive aggregate
increment over Native8 at both caps. The corresponding mean times for one
complete workload pass are 125.060 versus 91.632 seconds at $S=1$, and
83.154 versus 77.553 seconds at $S=8$. The ratios of totals are .733 and .933;
the medians of the 63 paired chunk totals are .733 and .934. Three of the
126 individual chunk/order measurements at $S=8$ favor Native8. After summing
both orders within each chunk, no chunk reverses at either cap. All of these
observations are retained; a positive aggregate does not imply that every
measured call is faster.

Generation alone takes 205.242 versus 138.392 seconds at $S=1$ and 114.205
versus 102.709 seconds at $S=8$, again summed over both passes. Model and
engine initialization, including compilation/capture where applicable, takes
66.469 and 65.307 seconds for the two processes. The two warmups together
take 2.186 and .956 seconds of service. Measured diagnostic serialization
takes 124.202 and 133.989 seconds across both arms; source tokenization
qualification takes 207.517 and 238.709 seconds. Complete worker wall times
are 866.515 and 820.853 seconds. The single allocation consumes 1,697 GPU
seconds, calculated from one allocated GPU times terminal main-job elapsed
time, including all startup, checks, diagnostics, shutdown, and idle overhead.
These costs are distinct from resident batch-service ratios. The experiment
does not measure online single-request latency, concurrent-client throughput,
end-to-end application savings, or proportional GPU billing.

\subsection{Actual decisions and descriptive quality}

All measured outputs are valid, but optimized execution is not strictly
decision-invariant. Native8 and Prefill1 differ in 12 of 4,030 paired decisions
at $S=1$ and 13 of 4,030 at $S=8$. At $S=1$, both arms repeat identically.
At $S=8$, the two repetitions differ on six of 2,015 inputs for Native8 and
one for Prefill1. Across scheduler caps, Native8 differs in 18 of 4,030
matched decisions and Prefill1 in nine; generated-ID differences have the
same counts. These records preclude a claim of exact finite-precision
trajectory equivalence.

After observing these differences, we recorded and performed a descriptive
CPU-only join of the actual timed alarms to the already exposed, released
code-agent references. No strategy or threshold was fitted, no new generation
was run, and all four cap/repetition conditions were retained. Each condition
contains the same 2,015 responses in 1,613 source components, with 1,014
positive and 1,001 negative references. This post hoc check attaches quality
to the measured speed points; it is not a new independent quality confirmation.

\begin{table}[t]
\centering
\begin{tabular}{rrrrr}
\toprule
$S$ & Repeat & Native8 F1 & Prefill1 F1 & $\Delta$ \\
\midrule
1 & 0 & 83.642 & 83.519 & $-0.123$ \\
1 & 1 & 83.642 & 83.519 & $-0.123$ \\
8 & 0 & 83.475 & 83.748 & $+0.273$ \\
8 & 1 & 83.598 & 83.704 & $+0.106$ \\
\bottomrule
\end{tabular}
\caption{Descriptive quality of the actual optimized outputs. F1 is in
percent; $\Delta$ is Prefill1 minus Native8 in percentage points. Repeat is
the counterbalanced order index, not a global Native-first/Prefill-first
condition. All four rows reuse the same references.}
\label{tab:optimized_full_quality_vseventeen}
\end{table}

For both $S=1$ repetitions, Native8 precision/recall is 90.286/77.909 and
Prefill1 is 90.263/77.712. At $S=8$, repeat 0 gives 90.160/77.712 versus
90.400/78.008, and repeat 1 gives 90.183/77.909 versus 90.297/78.008.
Prefill1 corrects versus introduces 2 versus 4 errors in each $S=1$
repetition, 6 versus 1 at $S=8$ repeat 0, and 4 versus 2 at repeat 1.
The F1 changes in Table~\ref{tab:optimized_full_quality_vseventeen} are small
but nonzero. Together with the full-workload service reduction, they support
a useful, configuration-dependent quality--cost point, rather than lossless
decoding or a general quality improvement. Complete aggregate phase costs,
confusion counts, all 126 paired timing observations per cap, and artifact
hashes are retained for audit.

\section{Artifact provenance and complete exploration costs}
\label{app:artifacts}

The preserved evidence manifest binds each completed scorer output to its SHA-256 digest, source/protocol identities, and actual allocation cost. It separates the prospective readout study from later exposed-cohort controls and replay. Original execution traces remain in the preserved experiment archives. The aggregate records retain failure costs, adverse outcomes, and correction lineage; they do not replace raw traces for independent rerunning.

The project ledger records 90,659 GPU-seconds (25.1831 GPU-hours) across 121 terminal allocations, including abandoned directions, loading, failures and idle/CPU work. This includes a 281-second qualification/timing pilot outside the labeled quality populations. The six-arm DEV assay, eleven-arm modern confirmation, earlier physical cascade and evidence/balanced-runtime assay contribute 479, 1,907, 999 and 593 seconds. The later binary/compiled controls and Boolean-grammar repair contribute 577 and 167 seconds; original and repaired results remain. The complete optimized workload adds 1,697 seconds; 747 seconds of auxiliary interface analysis outside this paper also remain counted. Each allocation is charged once. The complete ledger is preserved separately as \path{exploration_costs.json}. API expenditure is zero. Total exploration is distinct from method service and does not establish provider-price or energy savings.

\end{document}